\documentclass[runningheads]{llncs}

\usepackage{eccv}

\usepackage{graphicx}
\usepackage{amsmath}
\usepackage{amssymb}
\usepackage{booktabs}
\usepackage{multirow}
\usepackage{colortbl}
\usepackage{xcolor}
\usepackage{paralist}  %
\usepackage{pifont}  %
\usepackage{float}
\usepackage{tabularx}
\newcommand{\datasetname}{WorldPose\xspace}
\newcommand{\methodname}{our pipeline\xspace}

\newcommand{\xmark}{\ding{55}}%

\usepackage{wasysym}

\newcommand{\vectr}[1]{\boldsymbol{#1}}

\newcommand{\figref}[1]{\cref{#1}}
\newcommand{\tabref}[1]{\cref{#1}}
\newcommand{\secref}[1]{\cref{#1}}

\newcommand{\suppmat}{Supp. Mat\xspace}

\newcommand{\greentick}{\textcolor{cyan}{\checkmark}}
\newcommand{\redcross}{\textcolor{purple}{\xmark}}

\renewcommand{\paragraph}[1]{\noindent\textbf{#1}}

\let\titleold\title
\renewcommand{\title}[1]{\titleold{#1}\newcommand{\thetitle}{#1}}
\def\maketitlesupplementary
   {
   \newpage
        {\centering
        \Large
        \textbf{\thetitle}\\
        \vspace{0.5em}Supplementary Material \\
        \vspace{1.0em}}
   }

\usepackage{eccvabbrv}

\usepackage{graphicx}
\usepackage{booktabs}

\usepackage[accsupp]{axessibility}  %

\usepackage{hyperref}

\usepackage{orcidlink}

\begin{document}

\title{WorldPose: A World Cup Dataset for Global 3D Human Pose Estimation} 

\titlerunning{WorldPose}

\author{Tianjian Jiang \inst{*1} \and
Johsan Billingham\inst{*2} \and
Sebastian Müksch\inst{1} \and
Juan Zarate\inst{1} \and
Nicolas Evans\inst{2} \and 
Martin R. Oswald\inst{3} \and 
Marc Pollefeys\inst{1,4} \and 
Otmar Hilliges \inst{1}\and
Manuel Kaufmann \inst{1}\and
Jie Song \inst{\dagger1} 
}

\authorrunning{T.~Jiang and J.~Billingham et al.}

\institute{ETH Z{\"u}rich 
\and
FIFA 
\and
University of Amsterdam
\and
Microsoft
\\
\vspace{.1cm}\href{https://eth-ait.github.io/WorldPoseDataset/}{https://eth-ait.github.io/WorldPoseDataset/}
}

\maketitle

\begin{center}
\captionsetup{type=figure}
\includegraphics[width=1\textwidth]{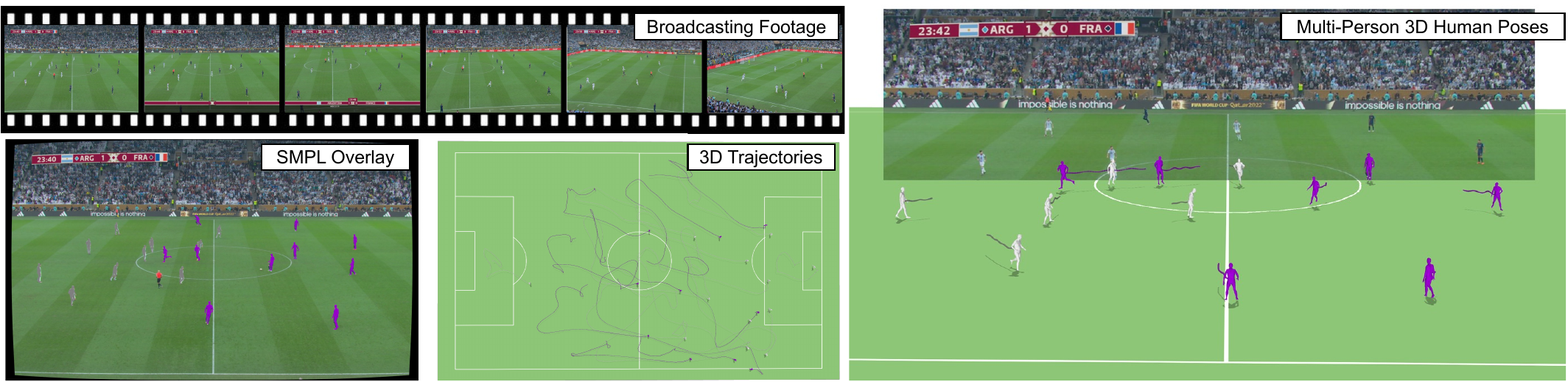}
\captionof{figure}{We leverage multi-view cameras to curate \datasetname, a comprehensive dataset designed for multi-person 3D human pose estimation with global trajectories.}
\label{fig:teaser}
\end{center}

\def\thefootnote{*}\footnotetext{These authors contributed equally to this work}
\def\thefootnote{\textdagger}\footnotetext{Now at HKUST(GZ)\&HKUST}

\begin{abstract}
We present \datasetname, a novel dataset for advancing research in multi-person global pose estimation in the wild, featuring footage from the 2022 FIFA World Cup. 
While previous datasets have primarily focused on local poses, often limited to a single person or in constrained, indoor settings, the infrastructure deployed for this sporting event allows access to multiple fixed and moving cameras in different stadiums.
We exploit the static multi-view setup of HD cameras to recover the 3D player poses and motions with unprecedented accuracy given capture areas of more than 1.75 acres (7k m$^2$).
We then leverage the captured players' motions and field markings to calibrate a moving broadcasting camera.
The resulting dataset comprises more than 80 sequences with approx 2.5 million 3D poses and a total traveling distance of over 120 km.
Subsequently, we conduct an in-depth analysis of the SOTA methods for global pose estimation.
Our experiments demonstrate that \datasetname challenges existing multi-person techniques, supporting the potential for new research in this area and others, such as sports analysis.
All pose annotations (in SMPL format), broadcasting camera parameters and footage will be released for academic research purposes. 


\end{abstract}
    
\section{Introduction}
\label{sec:intro}

The analysis of social interactions, crowd behavior, and team dynamics of human groups offers valuable insights for sociological research in various domains.
For instance, in sports science, precise 3D pose analysis for team activities could be a game-changing tool to improve training strategies, prevent injuries, and optimize overall team performance.
In recent years, significant advancements have been achieved in the development of Computer Vision and Deep Learning techniques for human pose estimation \cite{bogo2016keep,hmrKanazawa17,kolotouros2019learning,Li2021hybrik,li2022cliff,goel20234Dhumans,Sun2022BEV,Park2023TowardsRobustMultiPerson}.
However, existing pose estimation datasets used for the evaluation of these methods are predominantly designed for single individuals \cite{ionescu2014h36m,Mehta2017MPIINF3DHP,Hassan2019PROX,Zhang2022EgoBody,sigal2010humaneva,kaufmann2023emdb,Kazemi2013KTHMultiviewFootball}.
Those datasets that do feature multi-person scenarios are often constrained to lab-based settings \cite{joo2015panoptic,gemeren2016shakefive2,fieraru2020chi,guo2021expi,yin2023hi4d}, resulting in a limited number of individuals or restricted movement due to spatial constraints\cite{vonMarcard2018,Huang2022RICH,mehta2018mopots3d,khirodkar2023egohumans}.
However, many real-world scenarios involve large groups of people engaging in coordinated, dynamic activities that can occur in open, expansive outdoor areas and often involve moving cameras. Hence, these datasets are inadequate for capturing the complexity of these scenarios and especially insufficient for understanding the relative positions of multiple individuals over long trajectories.

Capturing a dataset that addresses these problems is a major challenge. 
First, the vast capture area renders marker-based methods relying on IR reflection impractical. Second, solutions based on body-worn sensors often encounter significant drifting issues, especially with dynamic body movements. 
Fortunately, due to the growing demand for sports analysis and Video Assistant Referee (VAR) systems, premier soccer events are often equipped with multi-view static camera systems, making it viable to apply multi-view markerless optical-based methods.

Hence, we leverage the capture infrastructure deployed in the 2022 FIFA World Cup stadiums to create \textit{\datasetname}, a large-scale multi-person pose estimation dataset with accurate calibration of a moving broadcasting camera. 
To obtain the desired annotations from such a premier camera setup, we base our method on classic optical-based methods due to their proven robustness. However, achieving the highest possible accuracy in our setting still requires significant adaptation of said methods. This is because: 1) The distance between the cameras and the subjects is large (to ensure comprehensive coverage of the field) and movement in soccer games is fast-paced with frequent occlusions, which results in even state-of-the-art (SOTA) models experiencing a notable decline in accuracy. 2) Calibrating the moving cameras (in our case the broadcasting cameras) remains a challenge due to the rapid movement of the camera and the limited distinctive features on soccer pitches.

To address these challenges, we carefully design a data acquisition pipeline which can be summarized in the following 3 steps:

\paragraph{Static Camera Calibration}
We first calibrate the static cameras
by initially treating the soccer field as a planar surface to compute the 2D homography between the image and the field plane. Then, we use the obtained homography as an initialization to solve a non-linear optimization that determines the camera parameters (including lens distortion) and accounts for the non-planar field. Finally, an additional photometric refinement process enhances the accuracy of the camera parameters to achieve pixel-level precision.

\paragraph{3D Human Pose and Shape Estimation}
Following static camera calibration, we then estimate the 3D pose and SMPL \cite{loper2015smpl} parameters in the world coordinate frame. The process starts by detecting and tracking each player's 2D keypoints. Due to the low resolution of the players in the image, we finetune the SOTA 2D detection and keypoints estimation models and leverage domain-specific knowledge to constrain the tracking algorithms. This process also includes a thorough manual review of the 2D detections with corrections if necessary. Since the static cameras are calibrated in the preceding step, we can triangulate these 2D keypoints to obtain global 3D joint coordinates. Subsequently, we fit SMPL parameters using 3D keypoint supervision, smoothness constraints, and a shape prior loss for improved accuracy.

\paragraph{Broadcasting Camera Calibration}
We first initialize the broadcasting camera parameters in a semi-automatic manner using a commercial software. However, in practice the software requires a pre-game scan by the camera operator, which is not available to us. To compensate, we leverage the 3D poses obtained from the previous step as an additional constraint, alongside the 2D field markings extracted with the software. Incorporating these additional constraints effectively enhances the accuracy and smoothness of the broadcasting calibration.

With this pipeline at hand, we curate a large-scale dataset that contains approximately 2.5 million accurate 3D human pose annotations, including global player trajectories disentangled from the camera's movement, which total a travelled distance of more than 120 km. 
When evaluated against Vicon\cite{Vicon}, the data acquisition pipeline yields a remarkable average error per joint of 8 cm, measured across global coordinates in a soccer stadium.

In summary, in this paper we contribute
\begin{inparaenum}[1)]
    \item \datasetname, to the best of our knowledge the first comprehensive dataset offering large-scale multi-person 3D poses paired with calibrated moving cameras. \datasetname provides accurate 3D human pose annotations with global trajectories and accurate broadcasting camera calibrations;
    \item Extensive evaluations of the accuracy of \methodname as well as baseline results of SOTA methods when evaluated on \datasetname. 
\end{inparaenum}
Our dataset and evaluation benchmarks will be made available for research.

\section{Related Work}

\paragraph{3D Human Pose Estimation}
Monocular 3D human pose estimation was revolutionized with the emergence of SMPL \cite{loper2015smpl,pavlakos2019expressive} and more powerful Deep Learning architectures.
The dominant approach is to estimate SMPL pose and shape parameters in camera-relative coordinates with a weak-perspective camera model \cite{Li2021hybrik,hmrKanazawa17,bogo2016keep,lassner2017unite,li2022cliff,kocabas2019vibe,song2020human,pymaf2021,Cho2022FastMETRO,Iqbal2021KAMA3K}, whereby some works consider multi-person estimation \cite{Sun2022BEV,Kocabas2021PARE,Sun2021ROMP,Huang2023ReconstrucingGroups,Park2023TowardsRobustMultiPerson,Li2023CoordinateTransformer,Wen2023Crowd3D,Wei2022MPSNet}, sometimes with a focus on larger crowds in recent years \cite{Sun2022BEV,Wen2023Crowd3D,Huang2023ReconstrucingGroups,goel20234Dhumans}.
Another line of work leverages multi-view setups for multi-person pose estimation.
Numerous methods~\cite{belagiannis1, shape_aware, mvpose, 4Dassociation, quickpose, shape_aware} formulate this problem as cross-view matching and association. More recent learning-based approaches choose to directly regress 3D human pose in 3D space~\cite{voxelpose, faster_voxelpose, voxeltrack, tempo}.
Simultaneously, there has been a growing interest in the recovery of \emph{global} human poses and camera trajectories from a single moving camera~\cite{yuan2022glamr,henning2022BodySlam,ye2023decoupling,Sun2023TRACE,kocabas2024pace}. Notably, GLAMR \cite{yuan2022glamr} attempts to recover global trajectories from per-frame local poses. SLAHMR \cite{ye2023decoupling} expands on this and considers camera motions to place humans in the scenes. Other works add scene constraints to the optimization, \eg, via optical flow \cite{Sun2023TRACE} or extract background features \cite{kocabas2024pace,henning2022BodySlam}.
In summary, we see a clear trend in the field towards estimating 1) 3D poses of more than a handful of people and 2) with global trajectories. However, progress is severely hampered by a lack of real, in-the-wild 3D reference data.
Our dataset \datasetname fills this gap and presents a challenging new setting with multiple people acting in a coordinated way in expansive space observed from moving cameras.

\paragraph{3D Human Pose Datasets}
\definecolor{LightGray}{gray}{0.9}
\begin{table}[t]
\caption{
\textbf{Comparison to related datasets}. ``GlobalTraj'': whether poses are captured in global coordinates. ``\#Frames'': number of frames without counting multiple views. ``\#Subjects'': number of subjects per frame. ``\#Poses'': total number of poses. ``Camera'': S (static), M (moving), M+Z (moving + zooming). *: rendered dataset. 
}

\centering
\begin{tabular*}{\textwidth}{@{\extracolsep{\fill}}lcccccc}
\toprule
Dataset & In the wild & GlobalTraj & Camera & \#Subjects  & \#Frames & \#Poses \\
\midrule
KTH   \cite{Kazemi2013KTHMultiviewFootball} & \greentick & \greentick & S & 2 & 0.8k & 0.8k\\
Panoptic Studio  \cite{joo2015panoptic} & \redcross & \greentick & S & 1-8 & 594k & 1.5M\\
H3.6M \cite{ionescu2014h36m}    & \redcross  & \greentick & S  & 1 & 630k  & 630k\\
PROX \cite{Hassan2019PROX}      & \redcross  & \greentick & S  & 1-2 & 88k  &  89k\\
3DPW  \cite{vonMarcard2018}     & \greentick & \redcross  & M  & 1-2 &  53k  &  75k\\
EgoBody \cite{Zhang2022EgoBody} & \redcross  & \greentick & M  & 2   & 220k  & 440k\\
RICH \cite{Huang2022RICH}       & \redcross  & \greentick & S  & 1-2 & 83k  &  85k\\
EMDB \cite{kaufmann2023emdb}    & \greentick & \greentick & M  & 1   & 105k  & 105k\\
SLOPER4D \cite{dai2023sloper4d} & \greentick & \greentick & M  & 1   & 100k  & 100k\\
BEDLAM*  \cite{Black2023Bedlam} & \greentick & \greentick & M & 1-10 & 380k & 1M\\

\midrule
\datasetname (Ours) & \greentick & \greentick & M+Z & 16 & 150k & 2.5M \\
\bottomrule
\end{tabular*}
\label{tab:dataset_comparison}

\end{table}

With \datasetname we propose a dataset for monocular multi-person 3D human pose estimation, both in camera-relative and global coordinates.
We highlight key differences to existing datasets in \tabref{tab:dataset_comparison} and discuss them here. 
While a few datasets are sourced from body-worn sensors \cite{vonMarcard2018,kaufmann2023emdb, dai2023sloper4d} or synthetically \cite{zhu2020reconstructing, Black2023Bedlam}, most are acquired from multi-view camera rigs \cite{sigal2010humaneva, ionescu2014h36m,joo2015panoptic,Mehta2017MPIINF3DHP,fieraru2020chi,yin2023hi4d,khirodkar2023egohumans,Huang2022RICH,guo2021expi} like ours.
However, the majority of existing datasets only show 1-2 people per image.
The only datasets that contain more than a handful of subjects per image are the seminal CMU Panoptic \cite{joo2015panoptic} and the recent BEDLAM \cite{Black2023Bedlam} (8 or 10 subjects per image). The former is recorded in a small lab-based setting preventing dynamic captures and the latter is rendered synthetically with uncoordinated motions and no interactions.
A closely related dataset is KTH Multiview Football II \cite{Kazemi2013KTHMultiviewFootball}, which also features footage of soccer games.
However, the 3D portion of \cite{Kazemi2013KTHMultiviewFootball} is limited in size with only 800 time instances from 3 views and 2 players.
In contrast, we provide footage with 10-20 subjects per frame on average from 150k frames.
In summary, in the landscape of 3D human pose datasets, \datasetname takes up a unique space: it contains more than double the amount of subjects per image than the previous largest dataset \cite{Black2023Bedlam}, features accurate global trajectories in a large capture area with a moving camera that is accurately calibrated, and contains high-quality SMPL pose and shape fits that are accurately tracked.

\paragraph{Sport Analysis}
Several notable works have contributed to the analysis of athletic activities in the domain of sports-related Computer Vision. \cite{rematas2018soccer} explores novel approaches to player and soccer scene reconstruction. \cite{zhu2020reconstructing} focuses on the reconstruction of basketball players. Both studies primarily utilize synthetic data extracted from game engines. \cite{giancola2018soccernet, deliege2021soccernet} provide a comprehensive soccer dataset for action analysis, albeit without 3D human poses.
The domain of sports camera calibration has also been a subject of extensive study. Works such as \cite{chen2019sports,homayounfar2017sports,citraro2020real,sha2020end,puwein2011robust,puwein2012ptz,puwein2014camera} address challenges in achieving accurate camera calibration in dynamic sports environments.
Finally, the comprehensive overview paper~\cite{thomas2017computer} surveys the current landscape and potential future directions at the intersection of Computer Vision and Sports Analysis.
In this paper, we focus on providing a new, comprehensive dataset featuring multi-person poses aligned with a single moving camera recording professional soccer games.

\section{Data}
\begin{figure*}[t]
    \centering
    \includegraphics[width=1.0\textwidth]{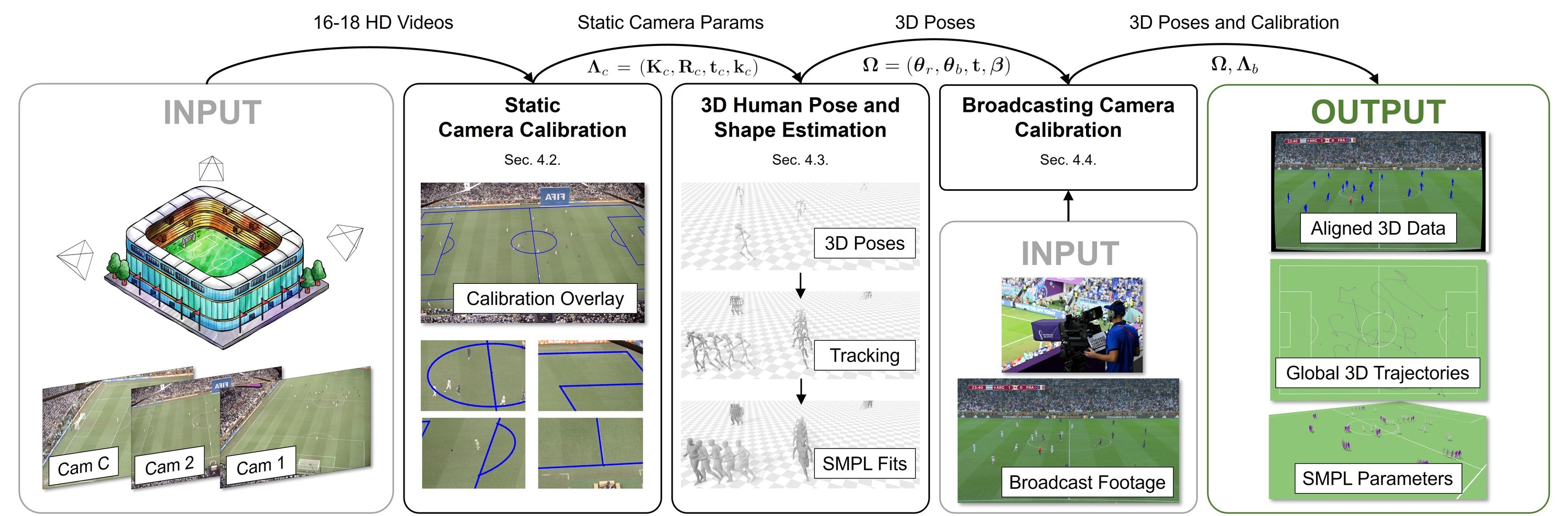}
    \caption{Method overview (from left to right): We take as input 16-18 high-resolution videos from statically placed cameras inside the stadium. The static cameras are calibrated by using hand-picked 2D points and photometric information (\secref{sec:static_cam_calibration}). This yields camera calibrations $\vectr{\Lambda}_c$ for every camera $c$, which we then use to triangulate and track 3D poses of each player (\secref{sec:pose_estimation}). We fit SMPL to the 3D pose data obtaining parameters $\vectr{\Omega}$. Finally, we calibrate a moving broadcasting camera to align the estimated 3D poses with broadcasted TV footage (\secref{sec:broadcast}). The method outputs 3D SMPL pose and shape parameters $\vectr{\Omega}$ of all soccer players, including their global trajectory, and accurate calibrations of the broadcast cameras $\vectr{\Lambda}_b$ with high-quality player pose reprojections. Stadium image sourced from \cite{FreePik}.}
    \label{fig:method}
\end{figure*}

We start by describing our dataset (\secref{sec:overview}), capture setup (\secref{sec:setup}) and notations (\secref{sec:notation}). The data acquisition pipeline is structured into three key components (see \figref{fig:method}): 1) calibrating static cameras around the stadium (\secref{sec:static_cam_calibration}), 2) estimating 3D human pose and SMPL parameters (\secref{sec:pose_estimation}), and 3) calibrating the moving broadcasting camera (\secref{sec:broadcast}).

\subsection{Data Overview}
\begin{figure}[t]
    \centering
    \includegraphics[width=0.9\linewidth]{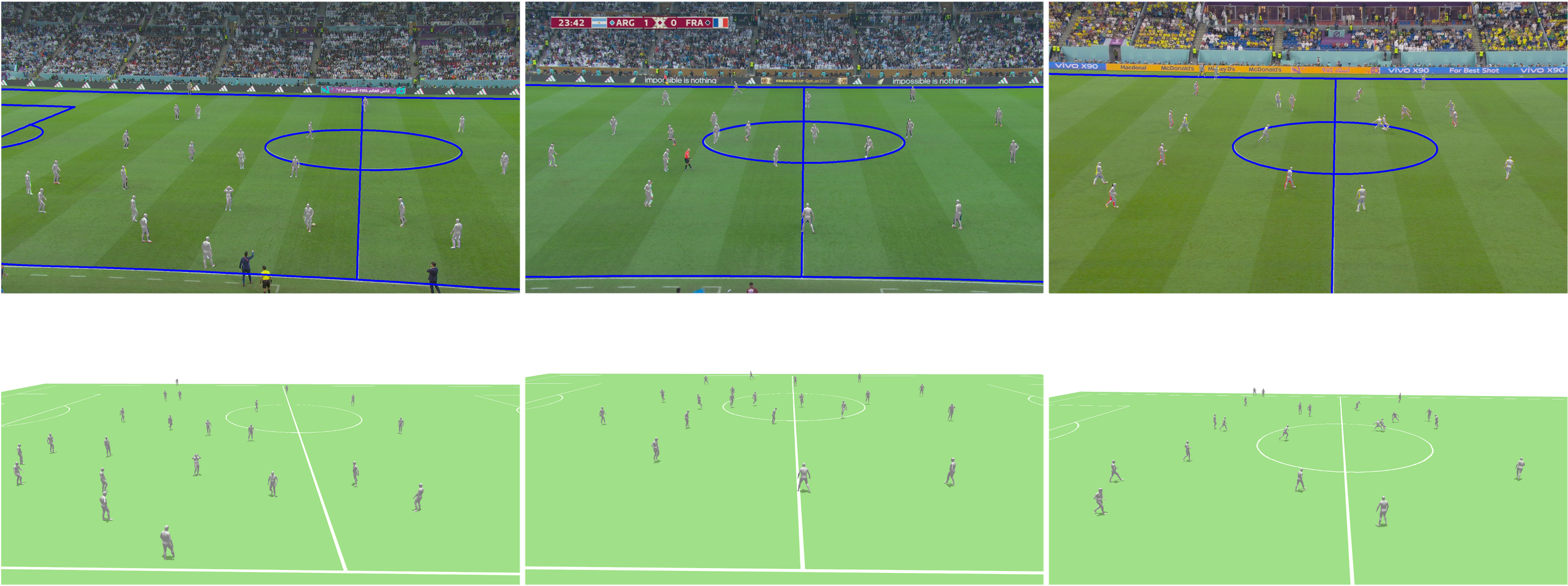}
    \caption{Sample images of the dataset. The first row displays the camera view and overlay, and the second row presents a novel 3D view to help readers understand the 3D locations of the subjects.}
    \label{fig:dataset}
\end{figure}

\label{sec:dataset}
\label{sec:overview}

We have collected a total of 88 broadcasting clips from the raw 1080p 50Hz TV program video footage of the quarter-finals and finals of the 2022 FIFA World Cup. Each clip is ensured to include at least one camera pan, such that the majority of the players in action will be captured.

The resulting dataset comprises 49.7 minutes of broadcasting footage and a total of 150k frames, containing 2.5 million recorded 3D poses in SMPL format. The total global distance travelled of all subjects amounts to more than 120 km. We present visualizations of our 3D data and reprojections in \figref{fig:teaser} and \figref{fig:dataset}. For additional visualization results and more statistics of the sequences, please refer to the supplementary document and video.

\subsection{Capture Setup}
\label{sec:setup}
The total capture space, equivalent to a standard soccer pitch measuring 105 $\times$ 68 meters, is covered by 16-18 strategically installed 1080p static cameras in each stadium (the number of cameras varies depending on the stadium). FIFA 3D LiDAR mappings of the World Cup stadium pitches are utilized in the static camera calibration process for better accuracy.

\subsection{Notations}
\label{sec:notation}
For each camera $c$ of the static multi-view cameras we denote the intrinsic parameters as $\mathbf{K}_c \in \mathbb{R}^{3 \times 3}$, the camera distortion coefficients as $\mathbf{k}_c \in \mathbb{R}^2$, and the extrinsic parameters involving camera rotation as $\mathbf{R}_c \in SO(3)$ and translation as $\mathbf{t}_c \in \mathbb{R}^3$. We summarize all camera parameters as $\vectr{\Lambda}_c = ( \mathbf{K}_c, \mathbf{R}_c, \mathbf{t}_c, \mathbf{k}_c)$.
Following static camera calibration, we discuss the estimation and tracking of 3D human poses $\mathbf{P} \in \mathbb{R}^{J \times 3}$ from those cameras. This process concludes with the registration of SMPL parameters, which cover shape parameters $\vectr{\beta} \in \mathbb{R}^{10}$, body pose parameters $\vectr{\theta}_b \in \mathbb{R}^{69}$, root orientation $\vectr{\theta}_r \in \mathbb{R}^{3}$, and translation $\mathbf{t} \in \mathbb{R}^{3}$ in world coordinates. These parameters are collectively denoted as $\vectr{\Omega} = (\vectr{\theta}_r, \vectr{\theta}_b, \mathbf{t}, \vectr{\beta})$.
Finally, we model the moving broadcasting camera with frame-wise focal lengths $f \in \mathbb{R}$, principal points $ \mathbf{c}_b \in \mathbb{R}^2$, radial distortion coefficients $\mathbf{k}_b \in \mathbb{R}^3$ and camera rotation $\mathbf{R}_b \in SO(3)$. We assume the camera location $\mathbf{C} \in \mathbb{R}^3$ in world space to remain constant across frames within each clip.
We summarize the broadcasting camera parameters as $\vectr{\Lambda}_b$.

\subsection{Static Camera Calibration}
\label{sec:static_cam_calibration}
For static camera calibration, we implement a multi-stage strategy inspired by the classic approach introduced in \cite{zhang2000flexible}.
In the first stage, we approximate the soccer field as a planar surface and estimate the 2D homography between the image and the plane. Building on this, in the second stage, we utilize the homography obtained as an initialization to solve a non-linear optimization problem, aiming to determine all camera parameters, including distortion.
This stage accounts for the fact that the pitch is not perfectly planar as a roughly 20 cm large field crown ensures water drainage.
Finally, in the third stage, an additional photometric refinement process is applied to further refine the camera parameters, ensuring pixel-level accuracy is achieved.

\paragraph{Stage 1}
We denote the 3D template of a soccer field as $\mathcal{S} = \{ \mathbf{X}_i'\}$, which consists of real-world pitch measurements $ \mathbf{X}_i' \in \mathbb{R}^3$ at characteristic field line markings obtained from official FIFA 3D LiDAR mappings of the World Cup stadium pitches.
Given this template, we can estimate camera parameters by aligning the image with the projections of $\mathcal{S}$. Initially, we project the 3D markings to a flat plane, denoted as $\mathbf{x}_i' \in \mathbb{R}^2$, and manually identify 2D correspondences in the image $\mathbf{x}_{i,c} \in \mathbb{R}^2$.
These 2D-to-2D correspondences are related via a 2D homography $\mathbf{x}'_{i} = \mathbf{H}_c \mathbf{x}_{i,c}$, whereby we can solve for $\mathbf{H}_c \in \mathbb{R}^{3 \times 3}$ using the direct linear transformation algorithm.
To determine the 9 parameters of $\mathbf{H}_c$ we select a few more than 4 correspondence pairs to make the problem over-determined.
Through decomposition of $\mathbf{H}_c$, we obtain camera parameters $\mathbf{K}_c [\mathbf{R}_c \mid \mathbf{t}_c]$.

\paragraph{Stage 2}
Considering that
\begin{inparaenum}[1)]
    \item the soccer pitch is not a flat plane in reality, and
    \item the output of the last stage does not account for camera distortion,
\end{inparaenum}
an additional non-linear optimization process is employed to improve the estimation.
To do so, we find the closest 3D point $\mathbf{X}'_i$ for every $\mathbf{x}_{i,c}$ and then refine the camera parameters with these 3D-to-2D correspondences:
\begin{equation}
\vectr{\Lambda}^\ast_c \in \arg\min_{\vectr{\Lambda}_c} \sum_i \|\Pi(\mathbf{X}'_i; \vectr{\Lambda}_c) - \mathbf{x}_{i,c}\|_2^2
\end{equation}
where $\Pi(\cdot; \vectr{\Lambda}_c)$ is a non-linear function that perspectively projects 3D points into the camera $c$ with distortion.

\paragraph{Stage 3}
For some views, where only one or two field corners are visible, the 3D-to-2D correspondences are too sparse to provide sufficient supervision.
Similar to \cite{rematas2018soccer}, we detect field lines in the image to obtain a denser set of correspondences. More specifically, we extract edge pixels using a line detector and construct a distance map the size of the image $\mathbf{D} \in \mathbb{R}^{H \times W}$, which for each pixel stores the distance to the nearest line pixels.
We then sample new 3D points from the field lines in the 3D template $\mathcal{S}$, project them into the image and minimize the distance of the projected point to the closest line pixel via a look-up in $\mathbf{D}$:
\begin{equation}
\vectr{\Lambda}^\ast_c \in \arg\min_{\vectr{\Lambda}_c} \sum_{\mathbf{X}' \sim \mathcal{S}} \mathbf{D}\left[\Pi \left(\mathbf{X}'; \vectr{\Lambda}_c\right)\right]
\end{equation}
where the operator $\mathbf{D}\left[(u, v)\right]$ indexes into the matrix $\mathbf{D}$ by rounding the projected point $(u, v)$ to integers.

\subsection{3D Human Pose and Shape Estimation}
\label{sec:pose_estimation}
With the static cameras calibrated, we now turn to estimating and tracking the 3D pose of each player, followed by fitting SMPL to the 3D pose.

\paragraph{Human 3D Pose Estimation and Tracking}
We initiate the process by detecting the bounding boxes of each player in each camera with ByteTrack \cite{zhang2022bytetrack}. Following this, we estimate the 2D poses with ViTPose \cite{xu2022vitpose}. As these models exhibit degraded performance on our low-resolution data, we employ a bootstrap approach to fine-tune them. %
Then we project the soccer field onto the images and eliminate all 2D detections located outside the field to filter out spurious detections in the audience.
Given the 2D keypoints $\{\mathbf{p}_{j,c}^t \mid j \in  (1, m) \}$ of the $j$-th player in frame $t$ of camera $c$ we triangulate the 3D pose denoted as $\mathbf{P}_{j}^t \in \mathbb{R}^{3J}$.
To track a player, we associate the 2D keypoint detections with 3D pose estimations from the previous frame $\{\mathbf{P}_i^{t-1} \mid i \in (1, n) \}$ using the following affinity function $A$:

\begin{equation}
A(\mathbf{P}_i^{t-1}, \mathbf{p}_{j,c}^t) = -\operatorname{PointToRayDist}\big(\mathbf{P}_i, \Pi^{-1}(\mathbf{p}_{j,c})\big)
\end{equation}
In other words, we compute the smallest distance of point $\mathbf{P}_i$ to the ray that results from the unprojection of $\mathbf{p}_{j,c}$ via $\Pi^{-1}$. The point-to-ray distance is averaged over the joints of the player.
We do this for all $i, j$ resulting in a $m \times n$ affinity matrix, so the tracking can be efficiently solved using a greedy matching algorithm.
When 3D poses from the previous frame are not available (the first frame or when a player track is lost), we utilize epipolar distance-based association to estimate new 3D poses from unmatched 2D poses.

\paragraph{Bundle Adjustment}
Although we have achieved good alignment for the field markings after the static camera calibration stage, the presence of inevitable measurement errors has motivated us to add a bundle adjustment stage to improve the 3D pose estimation accuracy.
To do so, we first hand-select a few frames per sequence where the 3D pose keypoints are of the highest quality, denoted as $\mathcal{P} = \{ \mathbf{P}_j^t \}$
Assume that all the 3D player joints in $\mathcal{P}$ and 3D field markings in $\mathcal{S}$ are merged into a set $\mathcal{X}$.
The bundle adjustment is then implemented by refining the camera parameters as follows:
\begin{equation}
\vectr{\Lambda}^\ast_c \in \arg\min_{\vectr{\Lambda}_c, \mathcal{P}} \sum_{\mathbf{X} \in \mathcal{X}} I_{j,c} \|\Pi(\mathbf{X}_j; \vectr{\Lambda}_c) - \mathbf{x}_j\|^2
\end{equation}
where $I_{j,c}$ indicates whether point $\mathbf{X}_j$ is visible in camera $c$ and $\mathbf{x}_j$ is the corresponding 2D point detection.

The bundle adjustment process also serves a valuable purpose by identifying outliers, where we check the points with large reprojection error after bundle adjustment and correct any mis-annotated points.%

\paragraph{SMPL Registration and Refinement}
Given the 3D poses of all players, we first estimate the SMPL shape, $\vectr{\beta} \in \mathbb{R}^{10}$ for each player as follows. We omit player subscripts for clarity. Assume $\mathcal{J}$ is the SMPL joint regressor, regressing 3D SMPL joints from mesh vertices.
Further, let $\ell$ be a function that extracts all bone lengths from a skeleton into a vector of size $\mathbb{R}^{J-1}$.
We can extract bone lengths from the SMPL template mesh $\bar{\mathbf{T}}$ and the shape blend shapes $\mathbf{B}_i$ and compare them to the bone lengths of $\mathbf{P}$ to directly estimate the SMPL shape $\vectr{\beta}$:
\begin{equation}
\vectr{\beta}^\ast = \arg\min_{\vectr{\beta}} \left|\ell(\mathbf{P}) - \ell\big(\mathcal{J}(\bar{\mathbf{T}})\big) - \sum_{i=1}^{10} \beta_i \ell\big(\mathcal{J}(\mathbf{B}_i)\big)\right|
\end{equation}

In the next step, we align the SMPL root location to $\mathbf{P}$ by minimizing the distance between the hips and torso keypoints of the SMPL model and the corresponding joints in $\mathbf{P}$.
To do so, we extract 3D keypoints from SMPL with a linear regression from the SMPL vertices, denoted as $\hat{\mathbf{P}} = \mathcal{J}(\mathcal{M}(\vectr{\Omega}))$. During this optimization, we freeze all SMPL parameters except translation and global orientation.

Finally, this output serves as an initialization to fit all the SMPL parameters $\vectr{\Omega} = (\vectr{\theta}_r, \vectr{\theta}_b, \mathbf{t}, \vectr{\beta})$ with several cost terms as defined in the following. First, we employ the 2D keypoint reprojection energy term on all joints, note for optmization of poses we align the projected 2D keypoint with the hip to minimize errors introduced by a misaligned root
\begin{equation}
E_\text{data} = \|\Pi(\hat{\mathbf{P}}) - \mathbf{P}_i\|^2_2
\end{equation}

Additionally, we ensure smoothness of motions for each trajectory and follow \cite{pavlakos2019expressive} to incorporate shape regularization via two losses:
\begin{equation}
E_\text{smooth} = \sum_t \|\hat{(\mathbf{P}}_{t+1} - \hat{\mathbf{P}}_{t}) - (\hat{\mathbf{P}}_{t} - \hat{\mathbf{P}}_{t-1})\|^2_2 \quad \quad E_\text{shape} = \|\vectr{\beta}\|^2_2
\end{equation}
With loss weights $\lambda_i \in\mathbb{R}_{\geq 0}$ the final loss to jointly refine all SMPL parameters is:
\begin{equation}
E_\text{refine} = \lambda_1 E_\text{data} + \lambda_2 E_\text{smooth} + \lambda_3 E_\text{shape}
\end{equation}

We also observed that initializing the SMPL body poses using estimations from broadcast footage empirically improves convergence speed and performance in challenging poses.

\subsection{Broadcasting Camera Calibration}
\label{sec:broadcast}

A standard broadcasting camera is employed to capture live FIFA World Cup games.
To calibrate it, we follow a similar strategy as for the static cameras whereby we first calibrate based on hand-picked 2D correspondences and then refine the estimation using 3D player information.
However, because the broadcasting camera is moving and operates under various levels of zoom, accurate calibration is a more challenging task than for the static cameras. Furthermore, given the size of our dataset manually picking 2D correspondences is in practice not a viable solution.
Thus, we use one of the leading commercial softwares built for semi-automatic calibration of broadcasting cameras \cite{VizArena} to facilitate an initial estimation.
The use of the software allows us to manually pick field markings in a few frames per clip and the rest of the clip is then tracked autonomously by the software.
In a next step, we then refine the calibrations with a dedicated optimization that ensures high-quality player and field marking reprojections, which we explain in the following.

We introduce two cost functions.
For field markings, we devise a 2D reprojection regularizer to ensure that the projection of the field markings after refinement does not deviate significantly from their initial positions:
\begin{equation}
E_{\text{field}} = \sum_{\mathbf{X}' \in \mathcal{X}} \rho \big(\Pi(\mathbf{X}'; \vectr{\Lambda}_b) - \mathbf{x}_{i,b}\big)    
\end{equation}
where $\rho$ is the Geman-McClure function \cite{Geman1987StatisticalMF} and $\mathbf{x}_{i,b}$ the corresponding detections in the broadcasting camera.
Additionally, we minimize the 2D reprojection loss between the 3D player keypoints and their corresponding 2D detections:
\begin{equation}
E_{\text{player}} = \sum_i \sum_j I_{i,j} \rho \big(\Pi(\hat{\mathbf{P}}_i; \vectr{\Lambda}_b) - \mathbf{p}_j\big)
\end{equation}
where $I_{i,j}$ is an indicator function that matches the 3D keypoints $\hat{\mathbf{P}}_i$ with the corresponding 2D keypoints $\mathbf{p}_j$.
To calculate $I_{i,j}$ we perform a weighted bipartite matching process with the following similarity function:
\begin{equation}
\operatorname{sim}(\mathbf{p}_i, \mathbf{p}_j) = \operatorname{sim}_{\text{IoU}}(\mathbf{p}_i, \mathbf{p}_j) \cdot \operatorname{sim}_{\text{bone}}(\mathbf{p}_i, \mathbf{p}_j)
\end{equation}
where for $\operatorname{sim}_{\text{IoU}}$, we calculate the intersection-over-union similarity for the bounding boxes of the players:
\begin{equation}
\operatorname{sim}_{\text{IoU}}(\mathbf{p}_i, \mathbf{p}_j) = \operatorname{IoU}\big(\operatorname{BBox}(\mathbf{p}_i), \operatorname{BBox}(\mathbf{p}_j)\big)
\end{equation}
and for $\operatorname{sim}_{\text{bone}}$, we calculate the mean cosine similarity between all the bones:
\begin{equation}
\operatorname{sim}_{\text{bone}}(\mathbf{p}_i, \mathbf{p}_j) = \frac{1}{J-1}\sum_{k=1}^{J-1} \operatorname{cos}(\mathbf{p}_{i,k}, \mathbf{p}_{j,k})
\end{equation}
With loss weights $\lambda_i \in\mathbb{R}_{\geq 0}$ the final objective function is then
\begin{equation}
    E_\text{calib} = \lambda_4 E_\text{field} + \lambda_5 E_\text{player}
\end{equation}
Implementation details are provided in the \suppmat.

\section{Experiments}
\subsection{Metrics}
We report 3 variants of the Mean Per Joint Position Error: 
\begin{inparaenum}[1)]
    \item \textbf{Global MPJPE (G-MPJPE)}, where we align the entire trajectories of all players between the prediction and ground-truth using a Procrustes Alignment (PA);
    \item \textbf{PA-MPJPE}: reporting the MPJPE error after aligning every player for every frame in both the prediction and the ground-truth using PA.
    \item For monocular multi-person global pose estimation, we additionally report the ratio between G-MPJPE and the corresponding length of the ground-truth trajectory, which we call \textbf{Per-Meter Drift}. It quantifies the deviation of the predicted trajectory from the ground-truth trajectory per meter.
\end{inparaenum}

\subsection{Comparison with Vicon}
\label{sec:vicon}

\begin{figure}[t]
    \centering
    \includegraphics[width=0.7\linewidth]{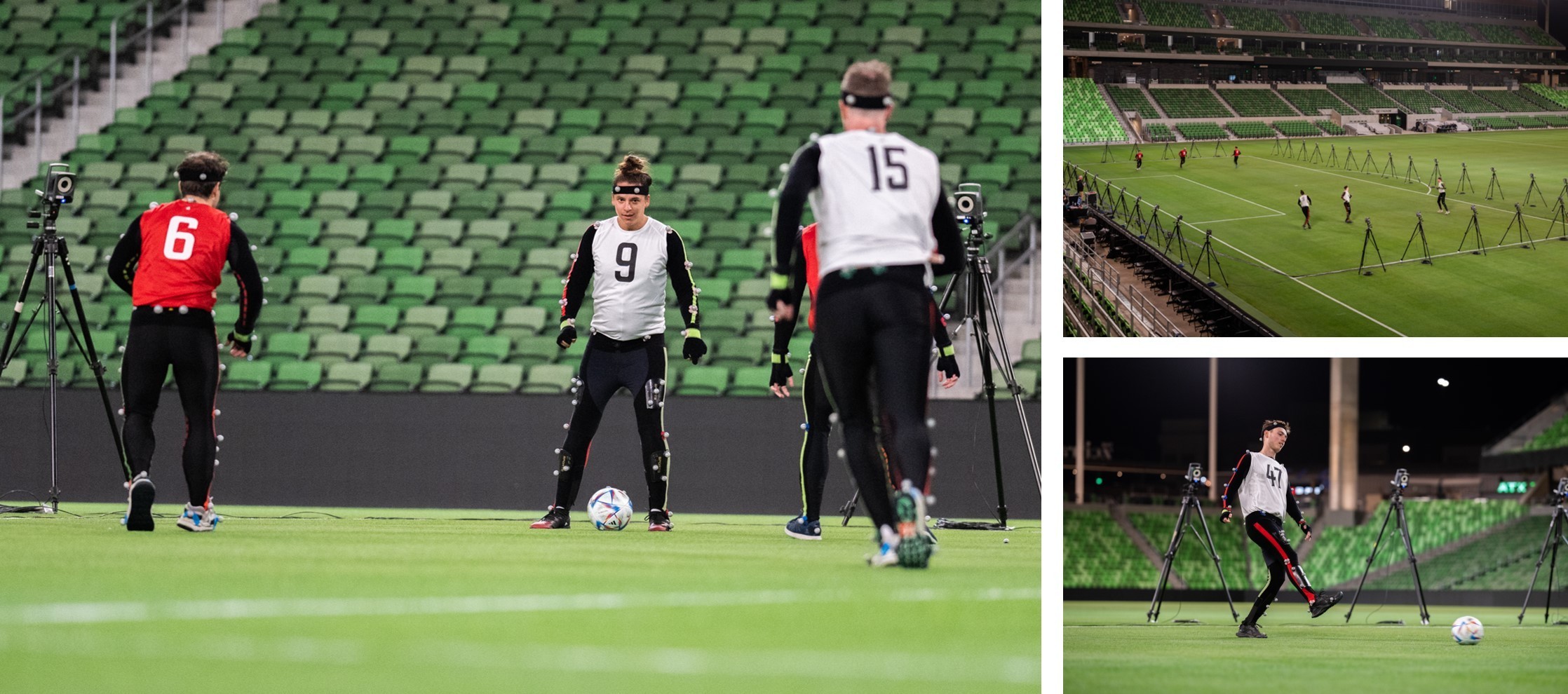}
    \caption{Vicon setup at night with 6 subjects playing in the penalty box.
    This data is used for evaluation purposes.}
    \label{fig:sevilla}
\end{figure}

\begin{table}
  \centering
  \caption{Evaluation of our pipeline on Vicon setup (\secref{sec:pose_estimation}). BA: Bundle Adjustment, $\mathcal{P}$: set of 3D player joints, $\mathcal{S}$: set of 3D field markings, +SMPL: SMPL refinement.}
  \resizebox{.50\linewidth}{!}{
  \begin{tabular}{lcc}
    \toprule
     & G-MPJPE [mm] $\downarrow$  & PA-MPJPE [mm] $\downarrow$ \\
    \midrule
    Base & 83.5 & 70.8 \\
    + BA ($\mathcal{S}$ and $\mathcal{P}$)  & 86.2 & 70.7 \\
    + BA ($\mathcal{S}$ only) & 548.4 & 75.4 \\
    + BA + SMPL  & \textbf{80.0} & \textbf{66.3} \\
    \bottomrule
  \end{tabular}
  }
  \label{tab:sevilla}
\end{table}

To evaluate the accuracy of our pipeline, we conduct trials with a setup featuring a Vicon \cite{Vicon} system to provide reference poses in a manageable portion of the pitch, \ie, the penalty box. 
Specifically, six players equipped with Vicon markers perform common motions, including dribbling, shooting the ball, and engaging in close body contacts, as depicted in \figref{fig:sevilla}. Additionally, 10 synchronized static cameras are deployed around the stadium and field measurements are conducted for calibration purposes.

We run the data through our pipeline, which also includes thorough manual review of the 2D detections and association with corrections if necessary. We then compare the output poses of our method to the poses supplied by the Vicon system.
We observe that the system was able to achieve a very low 6.6 cm error \wrt PA-MPJPE and 8.0 cm error \wrt G-MPJPE, which includes a measure of the global trajectory error. 
This underscores the high accuracy we can expect from \datasetname.

\begin{figure}[ht]
    \centering
    \includegraphics[width=0.8\linewidth]{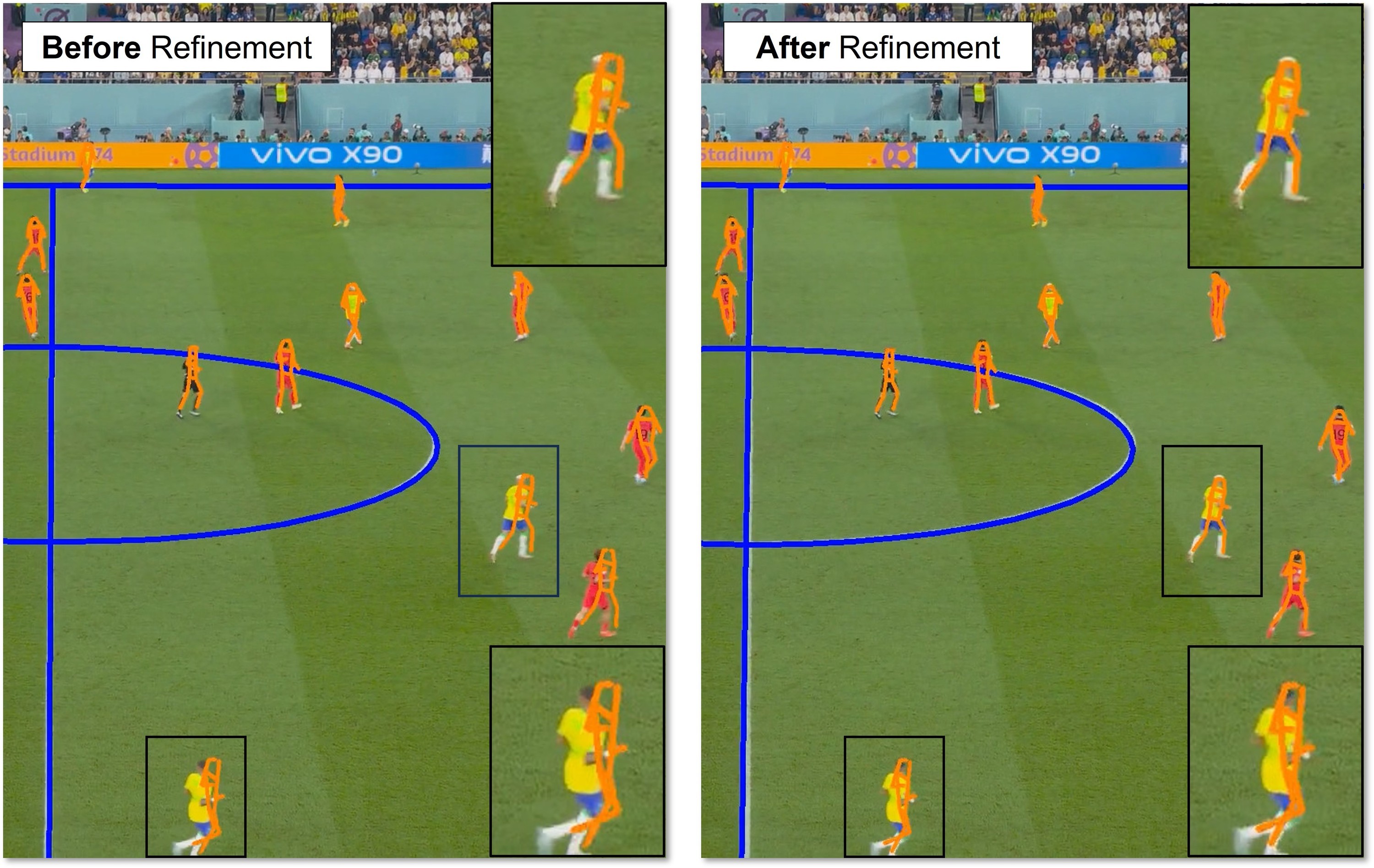}
    \caption{Visualization of broadcasting camera calibration before (left) and after (right) refinement with 3D poses (\secref{sec:broadcast}). Note the improved reprojections in the zoom-ins.}
    \label{fig:ablation-alignment}
\end{figure}

\subsection{Ablations}
\paragraph{Multi-View Human Pose Estimation} We also conducted ablation experiments to validate the design choices, as summarized in \tabref{tab:sevilla}. It shows that the incorporation of bundle adjustment and SMPL fitting leads to improved pose estimates (``+ BA + SMPL''). Furthermore, we performed an ablation study on the bundle adjustment process, optimizing cameras with respect to field markings alone (``+ BA ($\mathcal{S}$ only)'') and both field markings and keypoints  (``+ BA ($\mathcal{S}$ and $\mathcal{P}$)''). The results presented in \tabref{tab:sevilla} indicate that while both cases significantly reduce the reprojection error for field markings, the former tends to overfit to them, resulting in worse trajectory and pose estimations.  

\paragraph{Camera Calibration Ablation} We qualitatively compare the camera calibration for broadcasting footage before and after alignment in \figref{fig:ablation-alignment}. Similar to what was observed in  the bundle adjustment ablation study, a low reprojection error for field markings does not necessarily imply a low reprojection error for the players. In both subfigures, reprojection for field markings is highly accurate, but misaligned player reprojections are evident without our refinement.

\section{Benchmarks}
\label{sec:benchmark}

\begin{figure}[t]
    \centering
    \includegraphics[width=1.0\linewidth]{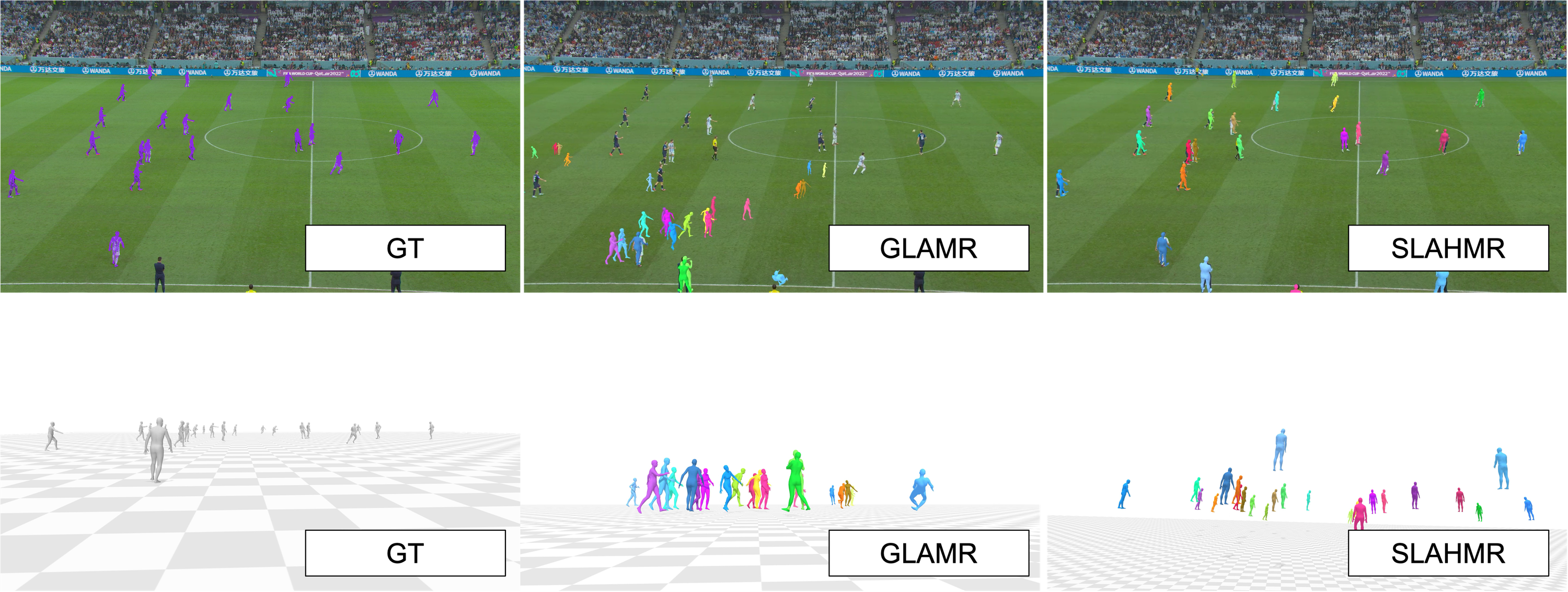}
    \caption{Visualization of global predictions from GLAMR and SLAHMR. The first row shows the camera view and the second shows the side view.}
    \label{fig:baselines}
\end{figure}

\begin{table}
  \centering
  \caption{\textbf{Results of state-of-the-art methods on \datasetname}. For ``per-person'' we estimate the optimal transformation for the trajectory of each player, otherwise we estimate a shared transformation for all the players.}
  \setlength{\tabcolsep}{3pt}
  \renewcommand{\arraystretch}{1.05}
  \resizebox{\linewidth}{!}{
  \begin{tabular}{lccc}
    \toprule
     & G-MPJPE [mm] $\downarrow$ & PA-MPJPE [mm] $\downarrow$ & Per-Meter Drift [cm/m] $\downarrow$\\
    \midrule
    Hybrik \cite{Li2021hybrik} & N/A &  78.8& N/A \\
    4DHuman \cite{goel20234Dhumans} & N/A & 116.5 & N/A \\
    \midrule
    GLAMR \cite{yuan2022glamr} & 18\,888.9 & 85.2 & 53.3 \\
    \bottomrule
    SLAHMR \cite{ye2023decoupling}                          & 8\,334.1 & 163.9 & 17.6 \\
    SLAHMR w/ GT Cameras            & 5\,837.2 & 199.6 & 10.7 \\
    SLAHMR w/o HuMoR  \cite{rempe2021humor}               & 9\,736.8 & 140.9 & 20.0 \\
    \bottomrule
    GLAMR  (per-person)               & 3\,749.7 & 85.2 & 8.3 \\
    SLAHMR (per-person)               & 4\,699.5 & 163.9 & 8.9 \\
    SLAHMR (per-person) w/ GT Cameras & 3\,818.5 & 199.6 & 7.2 \\
    \bottomrule
    
  \end{tabular}
  }
  \label{tab:evaluation}
\end{table}

We evaluate SOTA methods, GLAMR and SLAHMR on \datasetname. For GLAMR, we optimize the entire sequence simultaneously. For SLAHMR, as per the original paper, we first run DROID-SLAM \cite{teed2021droid} over the entire video, partition the video into chunks of 100 frames each and optimize each chunk separately. \tabref{tab:evaluation} and \figref{fig:baselines} summarize our key results discussed in the following:
\begin{enumerate}[1)]
\item  While GLAMR successfully aligns all subjects on the same plane, it struggles to generate reasonable global trajectories. The prediction of SLAHMR, on the other hand, is much closer to the ground-truth. However, despite us enforcing a single shared plane, SLAHMR still fails to align the subjects on that plane. 
\item To pinpoint the source of error in SLAHMR, we substitute the predicted camera parameters of DROID-SLAM with ground-truth values. This adjustment reduces the G-MPJPE and Per-Meter Drift by half. This observation indicates that DROID-SLAM encounters difficulties in accurately predicting camera parameters, which is unsurprising considering the pitch is nearly textureless and the background (the audience seats) are highly dynamic.
\item We note that both GLAMR and SLAHMR perform worse in terms of PA-MPJPE than the  method that they use for initialization (HybrIK for GLAMR, 4DHuman for SLAHMR).
\item We also note that leaving out HuMoR from SLAHMR (denoted as ``SLAHMR w/o HuMoR'') results in better performance. 
We hypothesize that this happens because HuMoR is conditioned on player height \wrt the plane, but the plane estimation is sometimes unreliable (see \figref{fig:baselines}).
\item We additionally report the ``per-person'' error, where we align the trajectory of each player individually with the ground-truth. Comparing this with the ``non per-person'' error, we observe a significant decrease in the evaluation metrics for both GLAMR and SLAHMR, regardless of whether ground-truth cameras are used. This suggests that a significant portion of G-MPJPE arises from incorrect relative positions between the players.
\end{enumerate}

With our experiments we show that while the current SOTA methods achieve impressive results in single person global human pose estimation, they: 
\begin{inparaenum}[1)]
\item encounter challenges when the area of movement expands,
\item have difficulty determining the relative positions between players, even when assuming a shared ground plane,
\item experience degraded performance when camera poses from the SLAM method are less reliable due to texture-less background or changing focal length.
\end{inparaenum}
We believe that providing a dataset featuring data in these challenging settings will facilitate exciting new research in this area.

\section{Conclusion}
In this paper we present \datasetname, a novel dataset that features high-quality 3D pose, shape, and global trajectory annotations of more than 10 subjects appearing simultaneously in monocular videos. %
With more than 2.5 Million poses, 88 total subjects, 150k frames, and 120 km travelled distance, \datasetname is a unique dataset contributing an important building block towards advancing multi-person pose estimation and motion modelling for real-world, coordinated interactions of large groups.
Our evaluations have shown that existing methods for global pose estimation struggle to produce convincing results. We hope that \datasetname will contribute to advancements of future methods in the field.

\paragraph{Limitations and Future work} A limitation of our method is its reliance on the quality of the 2D detections and the arrangement of the static cameras, and expensive manual interventions were required when a player was not adequately covered or when the image was blurry. Another limitation is that the data focuses on male events, resulting in an unequal gender representation among the participants. We aim to expand the dataset with future access to respective data. 

\section*{Acknowledgement}
This work was partially supported by the Swiss SERI Consolidation Grant "AI-PERCEIVE". We would like to express our gratitude to Vizrt, particularly Janick, Fabrizio, Jens, and Daniel, for their invaluable assistance in calibrating the broadcasting video. We also thank our other friends at Vizrt for their support. Computations were carried out in part on the ETH Euler Cluster.

\bibliographystyle{splncs04}
\bibliography{egbib}

\clearpage
\setcounter{page}{1}
\setcounter{section}{0}
\maketitlesupplementary

In this supplementary material, we provide additional information to complement the main text:
\begin{inparaenum}[1)]
\item more implementation details about the data creation (see \secref{sec:supp_implementation_details}),
\item more details about the baselines (see \secref{sec:supp_baselines}), and 
\item more statistics and sample images from \datasetname (see \secref{sec:supp_statistics}).
\end{inparaenum}

\section{Data creation}
\label{sec:supp_implementation_details}
\subsection{Static Camera Calibration}
\begin{figure}
    \centering
    \includegraphics[width=0.45\linewidth]{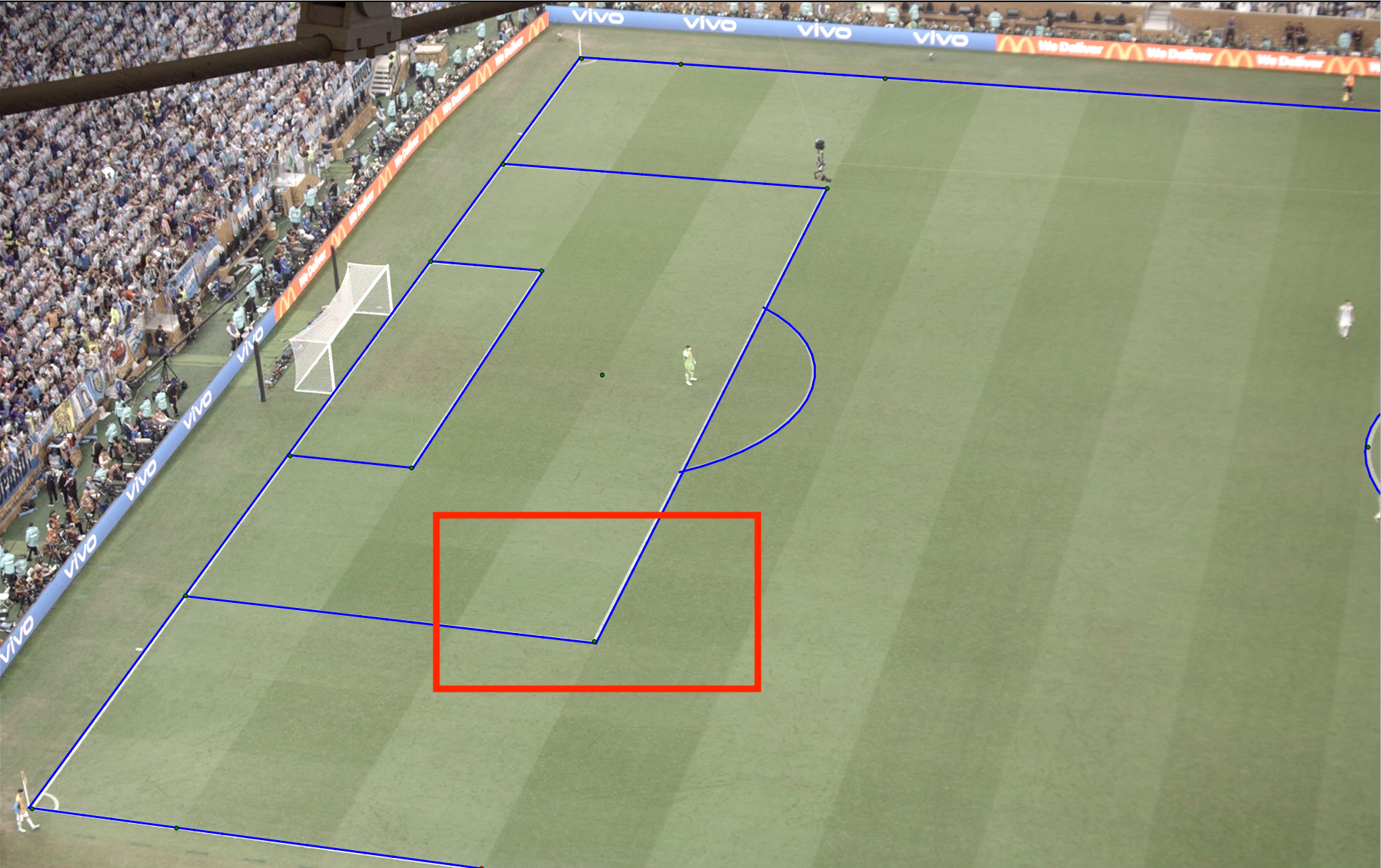}
    \includegraphics[width=0.45\linewidth]{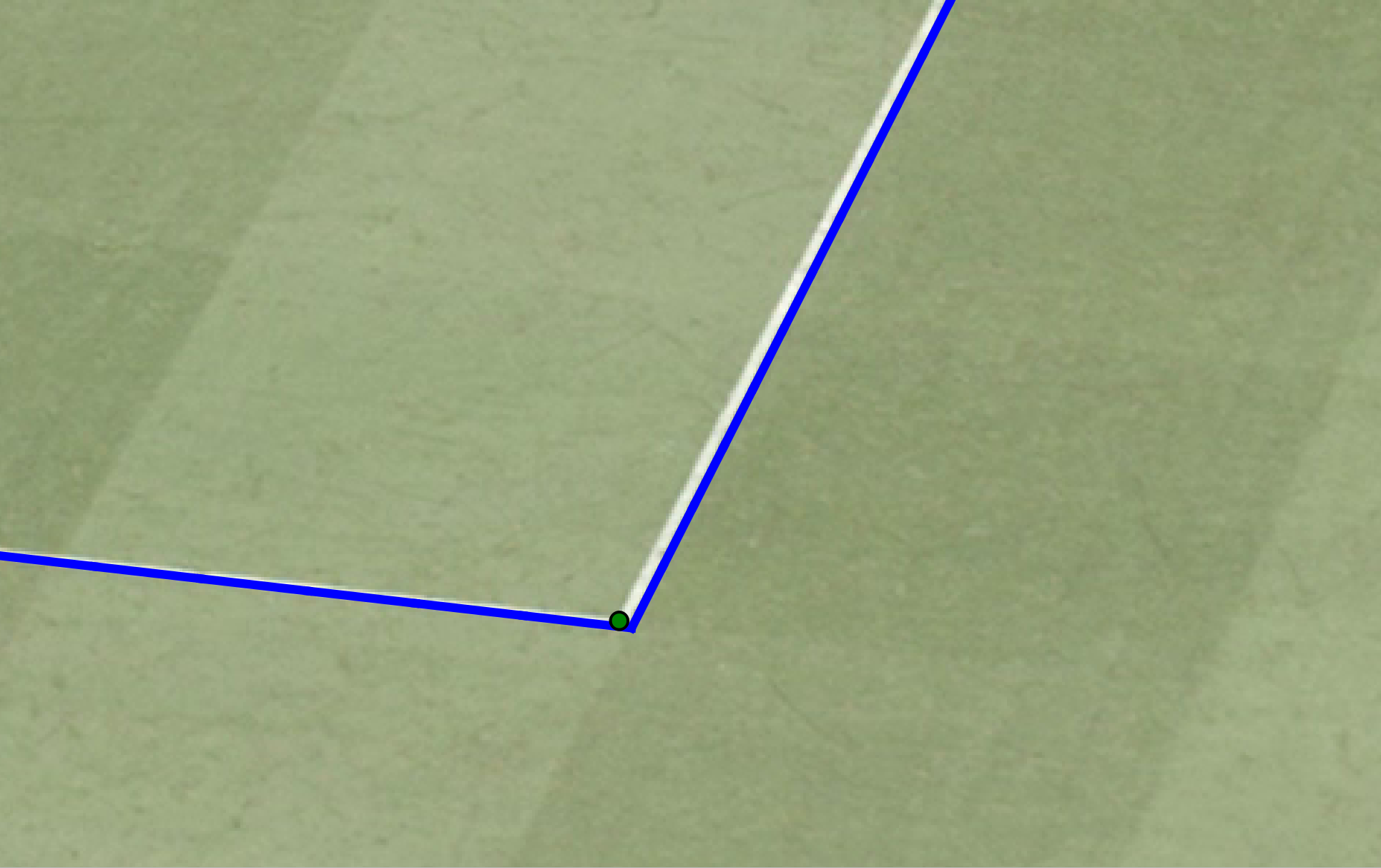}
    \caption{Visualization of the annotation tools: manual annotation (left) and zoomed-in view (right). The manually selected point is indicated by the green marker. The blue lines show the preview results of the projection using the manually selected points.}
    \label{fig:supp-annot-tool}
\end{figure}

For static camera calibrations, we developed a GUI program designed for the manual annotation of 2D points, as is shown in \figref{fig:supp-annot-tool}. This tool simplifies the process of adding and editing annotations. It also offers additional features, like
\begin{inparaenum}[1)]
    \item zooming, which is crucial for achieving pixel-level accuracy, and
    \item previewing, where it dynamically updates the estimation of camera parameters using the annotated 2d points and generates preview results of the projection of the field markings.
\end{inparaenum}

\begin{figure}
    \centering
    \includegraphics[width=0.45\linewidth]{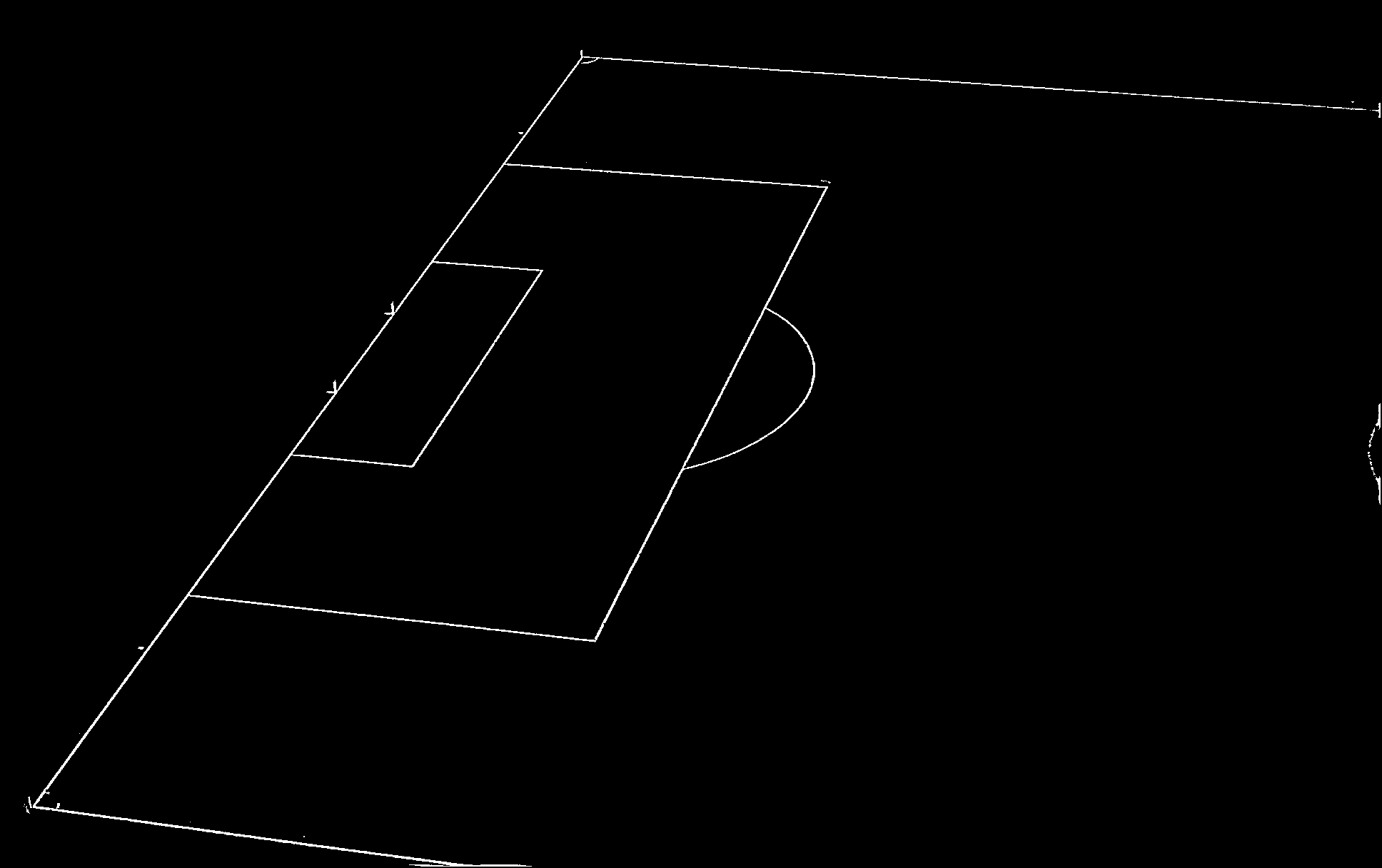}
    \includegraphics[width=0.45\linewidth]{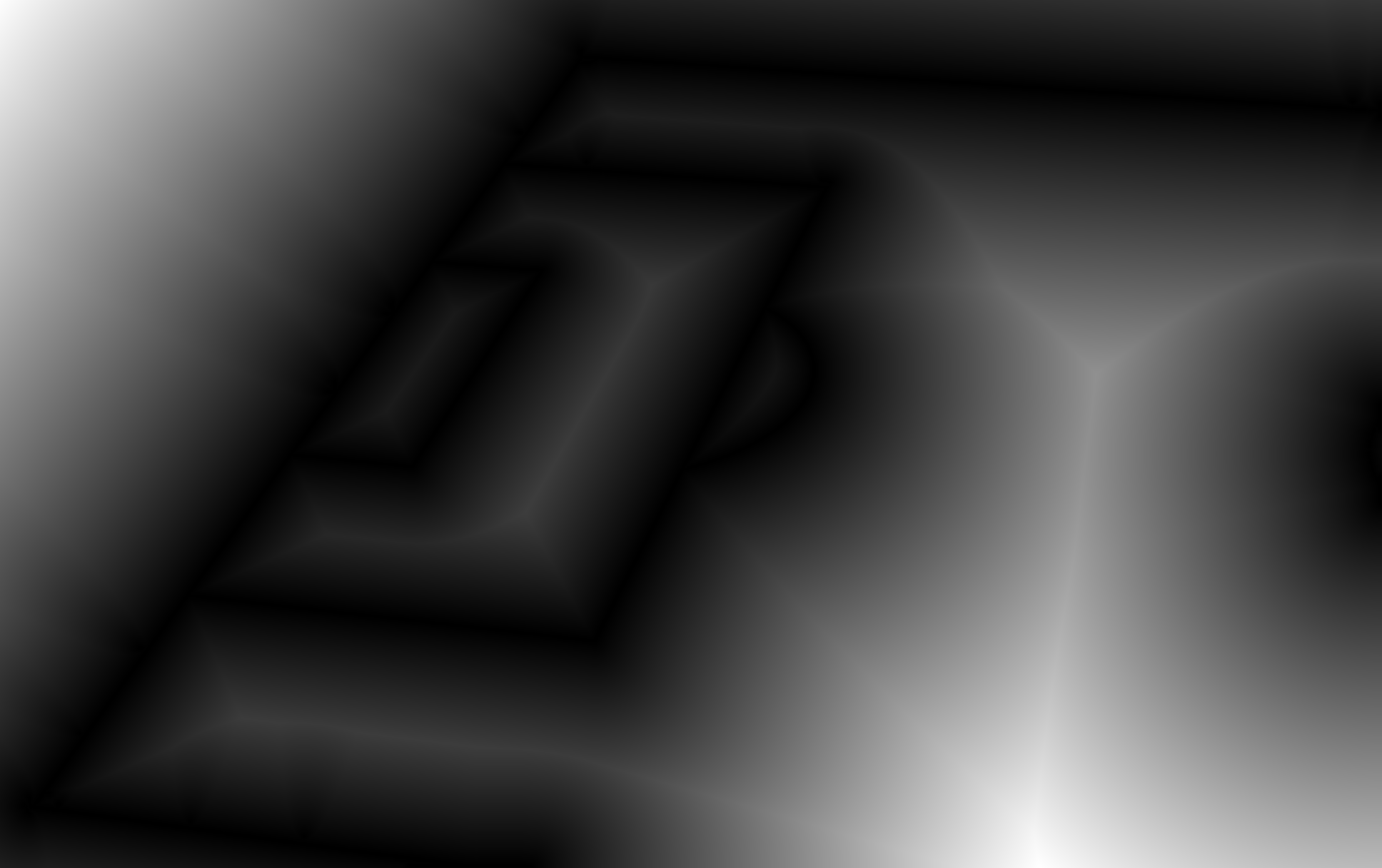}
    \caption{Visualization of the photometric refinement process: extracted field markings (left) and distance field induced from the field markings (right). The brightness corresponds to the distance from the field markings.}
    \label{fig:supp-annot-tool-lines}
    \vspace{-1cm}
\end{figure}

Subsequently, we employ the Canny detector \cite{canny1986computational} to extract the field markings from OpenCV \cite{bradski2000opencv}. The calibration results from the previous stage are also utilized to remove uninteresting lines. The detected lines are then converted into a distance field matrix, where each element of the matrix represents the distance to the nearest field markings. With the distance field we can further refine the camera parameters by minimizing the distance of the projected point of the field markings to the closest line pixel. The detected field markings and distance matrix are visualized in \figref{fig:supp-annot-tool-lines}.

\begin{figure}
    \centering
    \begin{subfigure}[b]{0.45\linewidth}
    \includegraphics[width=\linewidth]{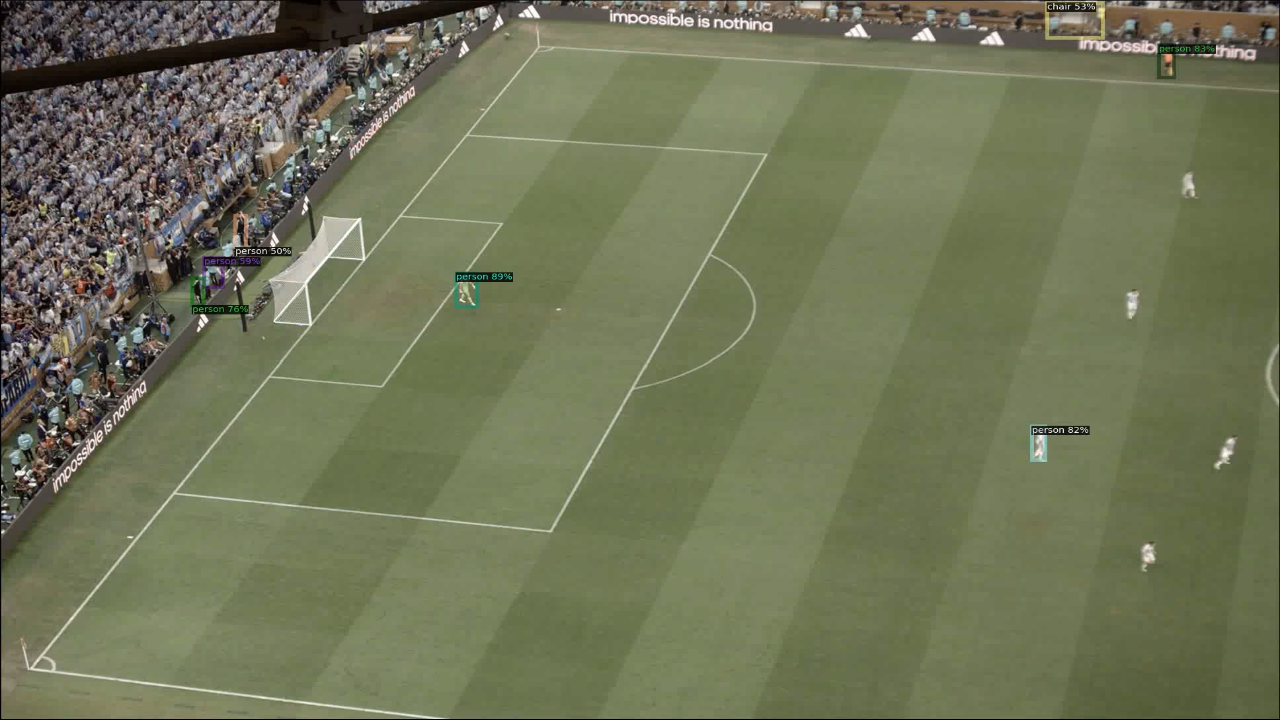}
    \includegraphics[width=\linewidth]{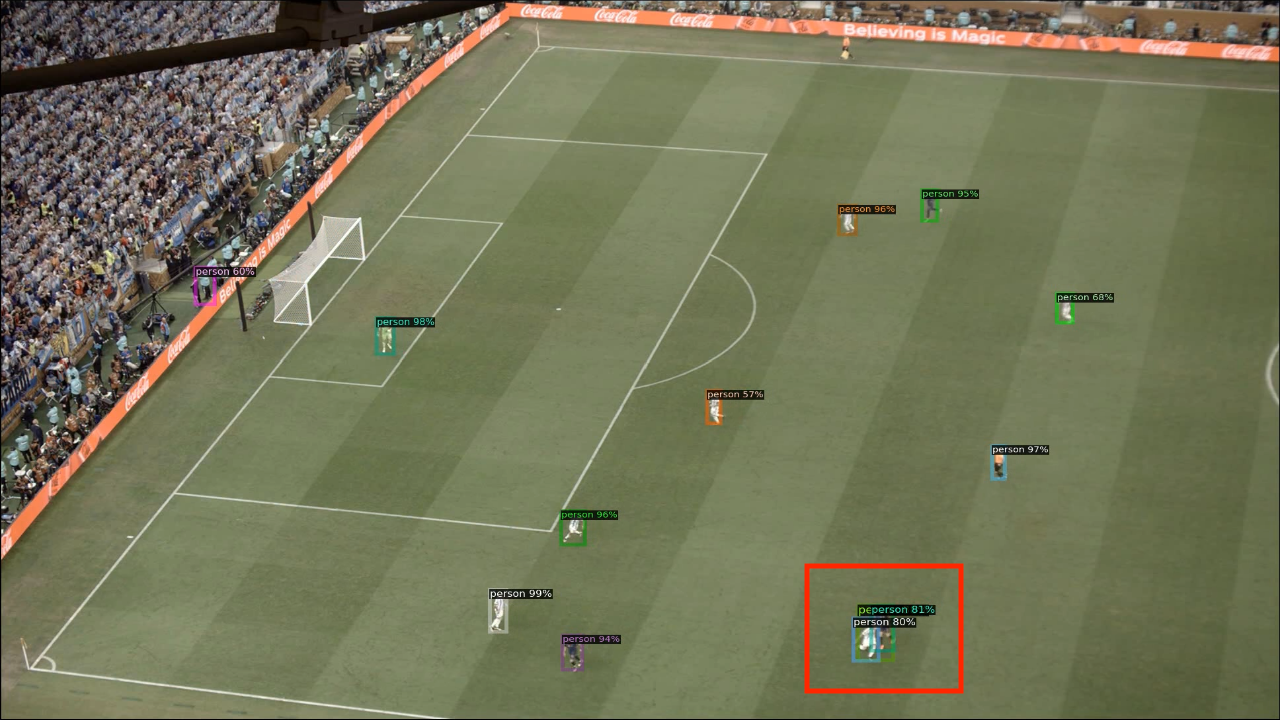}
    \caption{Results of SOTA models (Detectron2-R-101-FPN)}
    \end{subfigure}
    \begin{subfigure}[b]{0.45\linewidth}
    \includegraphics[width=\linewidth]{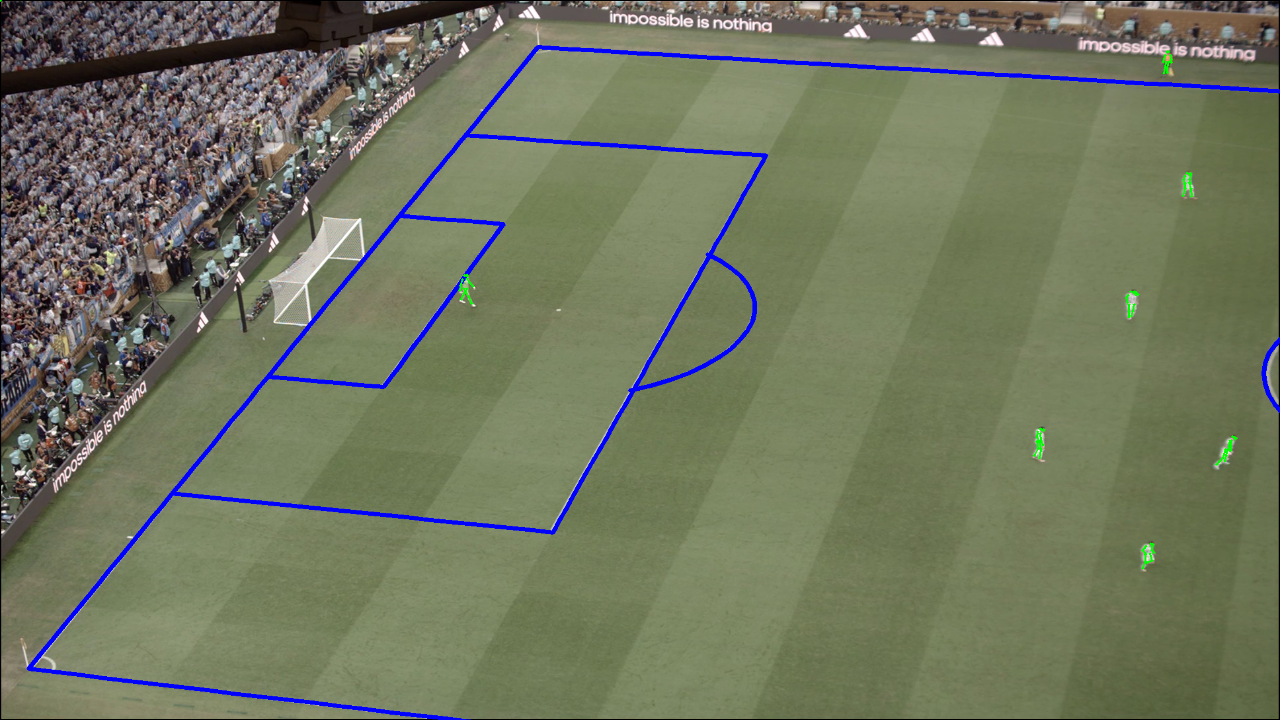}
    \includegraphics[width=\linewidth]{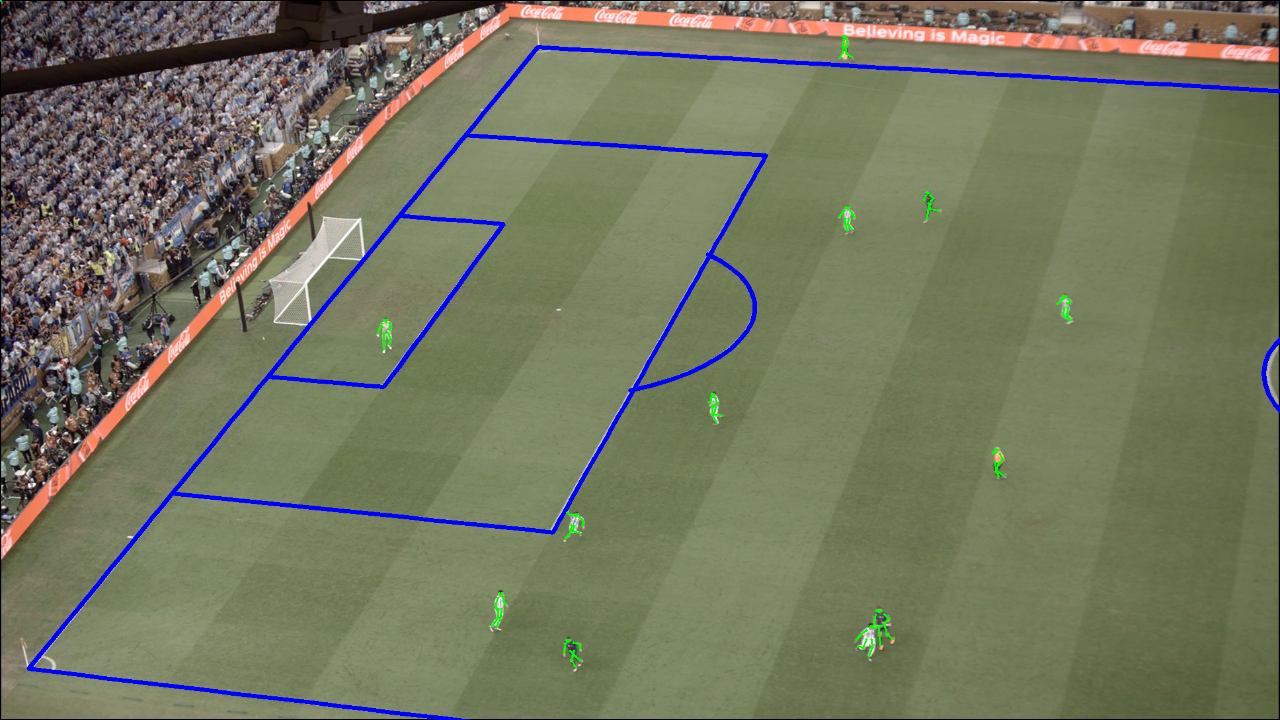}
    \caption{Results of ours\newline}
    \end{subfigure}
    \caption{Comparison of SOTA models vs. ours. Note the missing players in the top row and erroneous detections highlighted in the red rectangle (it detects 3 players instead of 2). We get rid of the uninteresting detections outside the field (top row, top of the image) by leveraging the camera calibration.}
    \label{fig:supp-detect}
\end{figure}

However, there are still several problems remained: 1) by default, we only estimate the $k_1$ and $k_2$ distortion coefficients of the cameras. However, this may not be sufficient in some cases, especially for side cameras with wide angles. These cameras need to be handled separately, and usually, adding the $k_3$ coefficient is sufficient to achieve relatively good results. 2) another issue is that even when the reprojection appears reasonable, it can be problematic for cameras looking at the penalty area, as shown in \figref{fig:supp-detect}. Due to the lack of corresponding points in the right half of the image, there can be multiple sets of parameters that provide roughly the same reprojection for the field lines but very different results for the players. Therefore, the keypoints of players must be considered in the calibration process. This means it often takes multiple iterations of the entire camera calibration and keypoint estimation process to achieve desired accuracy.

\subsection{Refinement of 2D detection results}
Due to the distance between the camera and the players, the resolution of the players is rather low, which will negatively impact the accuracy of 2D detections. Consequently, even SOTA models may frequently miss the players or produce erroneous detections, as illustrated in the figure below.

To address these issues, we initially ensemble the predictions of multiple SOTA detection models by concatenating their detections and running Non-Maximum Suppression to eliminate duplicate detection boxes. However, the ensemble model is relatively slow and occasionally produces incorrect detections. Therefore, we apply this slower method to a subset of broadcasting images. We manually inspect the results and remove incorrect detections. In this way, we semi-automatically annotate a small dataset and fine-tune the YOLO models with this dataset. Through this process, we achieve a 2D detection model with desired accuracy and speed.

\subsection{Estimating 3D Skeletons}
As described in Section 3.5, 3D pose can be acquired via triangulation. While this method performs well with accurate static calibration, small calibration errors—especially due to camera-subject distance—can lead to significant inaccuracies. In many cases, the back-projected rays from different cameras fail to converge at a single point. While bundle adjustment can improve results, challenges may persist in many scenarios. Therefore, we adopt a two-stage strategy: first, we estimate the 3D location of the mid-hips, and then during the optimization process, we normalize both the projected 3D and 2D keypoints:
$$
X^* = \arg\min_X \| \big(\Pi(X; \Lambda_c) - \Pi(X_{\text{midhip}}; \Lambda_c)\big) - \left(x - x_{\text{midhip}}\right) \|^2
$$
In essence, we first regress the mid-hips for each subject. Then, during 3D keypoint optimization, we align the projected 3D mid-hip with the mid-hip from the 2D detections, as we are primarily concerned with local poses.

\subsection{Broadcasting Camera Calibration}
For Broadcasting camera calibration we used Adam Optimizer\cite{kingma2014adam} with a learning rate of 10$^{-3}$ which empirically leads to slightly smoother camera parameters. Here, the hyperparameters are set to $\lambda_4=1$ and $\lambda_5=0.5$.

\begin{figure}
    \centering
    \includegraphics[width=0.7\linewidth]{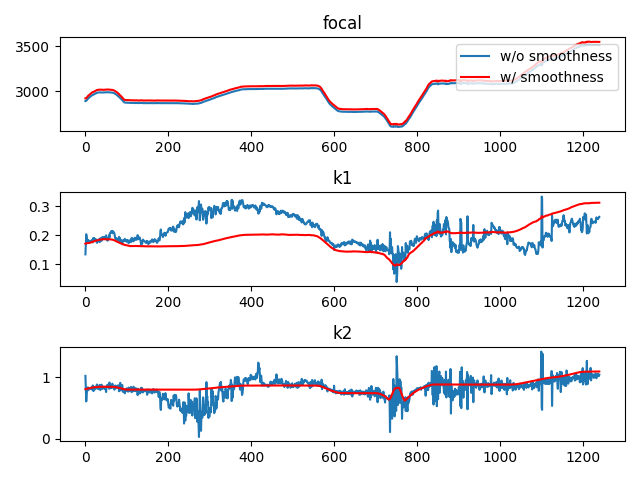}
    \caption{Visualization of the camera parameters: The X-axis represents the frame index, and the Y-axes represent the focal length, k1, and k2 respectively. The smoothness of all camera parameters (including distortion coefficients) improved with the additional smoothness regularizations.}
    \label{fig:supp-broadcast-params}
\end{figure}

However, while this may be sufficient for achieving a low reprojection error, it often leads to less smooth distortion coefficients, as demonstrated by the blue curves in \figref{fig:supp-broadcast-params}.

\begin{figure}
    \centering
    \includegraphics[width=0.8\linewidth]{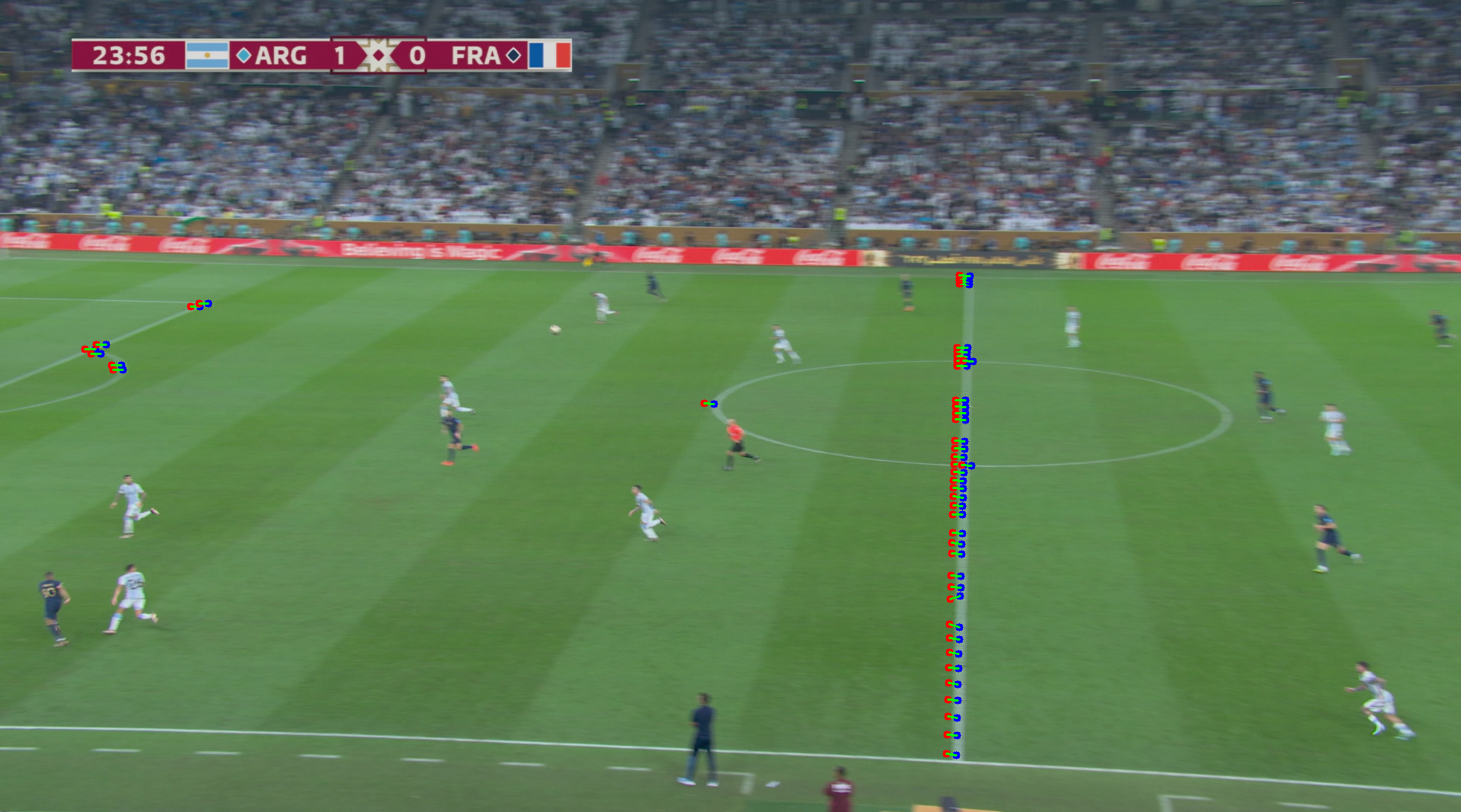}
    \caption{Visualization of optical flow regularization: We employ the iterative Lucas-Kanade method with pyramids \cite{bouguet2001pyramidal} to compute the optical flow for a sparse feature set consisting of points sampled from field markings. Here, the red points represent the sampled points from the previous frame, and the green points represent the predictions of the optical flow. Note outliers are removed with modified z-score.}
    \label{fig:supp-broadcast-optical-flow}
\end{figure}

To address this, we have also incorporated additional smoothness regularizers, including a camera smoothness term and optical flow regularization (see \figref{fig:supp-broadcast-optical-flow}). With these, we enforce that 1) the changes of focal length and distortion shall be smooth across frames, and 2) the reprojection of the keypoints and the optical flow prediction shall be close to each other. In this way, we can achieve not only accurate but also visually smooth reprojection. The weights of these two regularizers are adjusted subject to the clip after a manual check.

\subsection{SMPL Fitting}
One limitation of directly fitting SMPL from 3D poses is that the resulting poses lack prior knowledge of feasible 3D poses, which can lead to the reconstruction of unrealistic poses which happen to have low reprojection errors. As mentioned in the main paper, one solution is to initialize the SMPL body poses using estimates from broadcast footage, which typically have higher zoom levels and capture more detailed players. Thanks to the broadcast calibration process, we have associated players in the broadcast videos with 3D poses, and this enables us to leverage 2D-based SMPL estimators, such as \cite{goel20234Dhumans} or \cite{Li2021hybrik}, to initialize and regularize the body poses.

In practice, running the entire pipeline once is usually sufficient to achieve desirable results. However, while the association process (identifying the same players across cameras) is generally robust with adequate camera coverage, it can still fail in challenging scenarios, such as occlusions or difficult poses. One workaround is to rerun the pipeline multiple times, using the results from previous runs to initialize and filter out outlier detections. For example, we typically use the fitted SMPL parameters to filter erroneous 2D keypoint detections by comparing the projected 2D poses from SMPL with the estimated 2D poses using metrics such as cosine similarity. This approach in practice improves the robustness.

\section{Baseline Evaluation Details}
\label{sec:supp_baselines}
\subsection{GLAMR Baseline}
\paragraph{Implementation Details} For the GLAMR \cite{yuan2022glamr} baseline, we found that the official implementation was unable to detect many subjects within the frame. To address this we supply it with detection results generated by our preprocessing code that utilizes BYTETrack \cite{zhang2022bytetrack}. To achieve the best results, we run GLAMR on the entire video using a single A100 GPU.

\begin{figure*}[htbp]
\centering
\begin{subfigure}[b]{0.45\linewidth}
    \includegraphics[width=\linewidth]{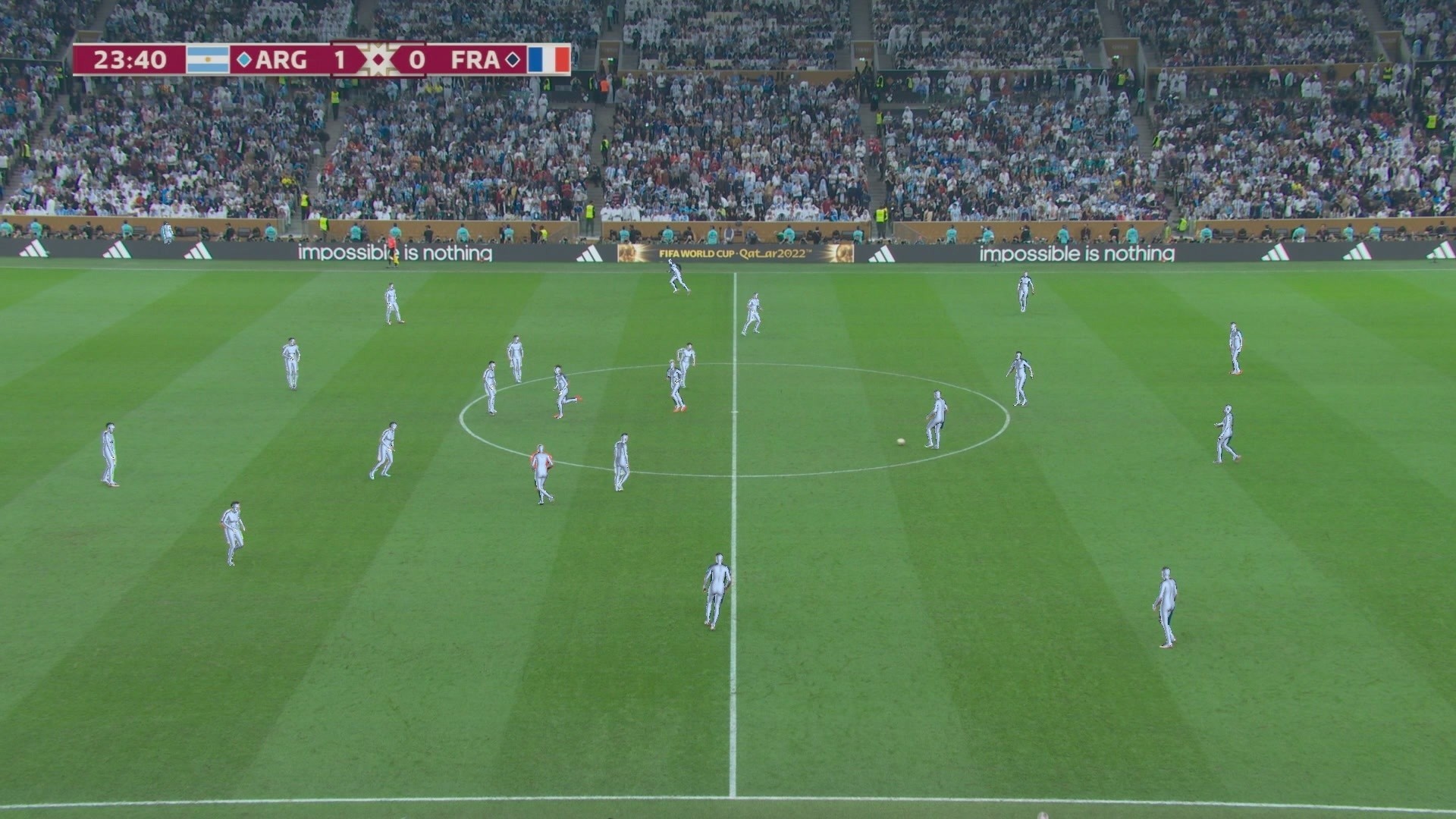}
    \caption{Hybrik (with our detections)}
\end{subfigure}
\begin{subfigure}[b]{0.45\linewidth}
    \includegraphics[width=\linewidth]{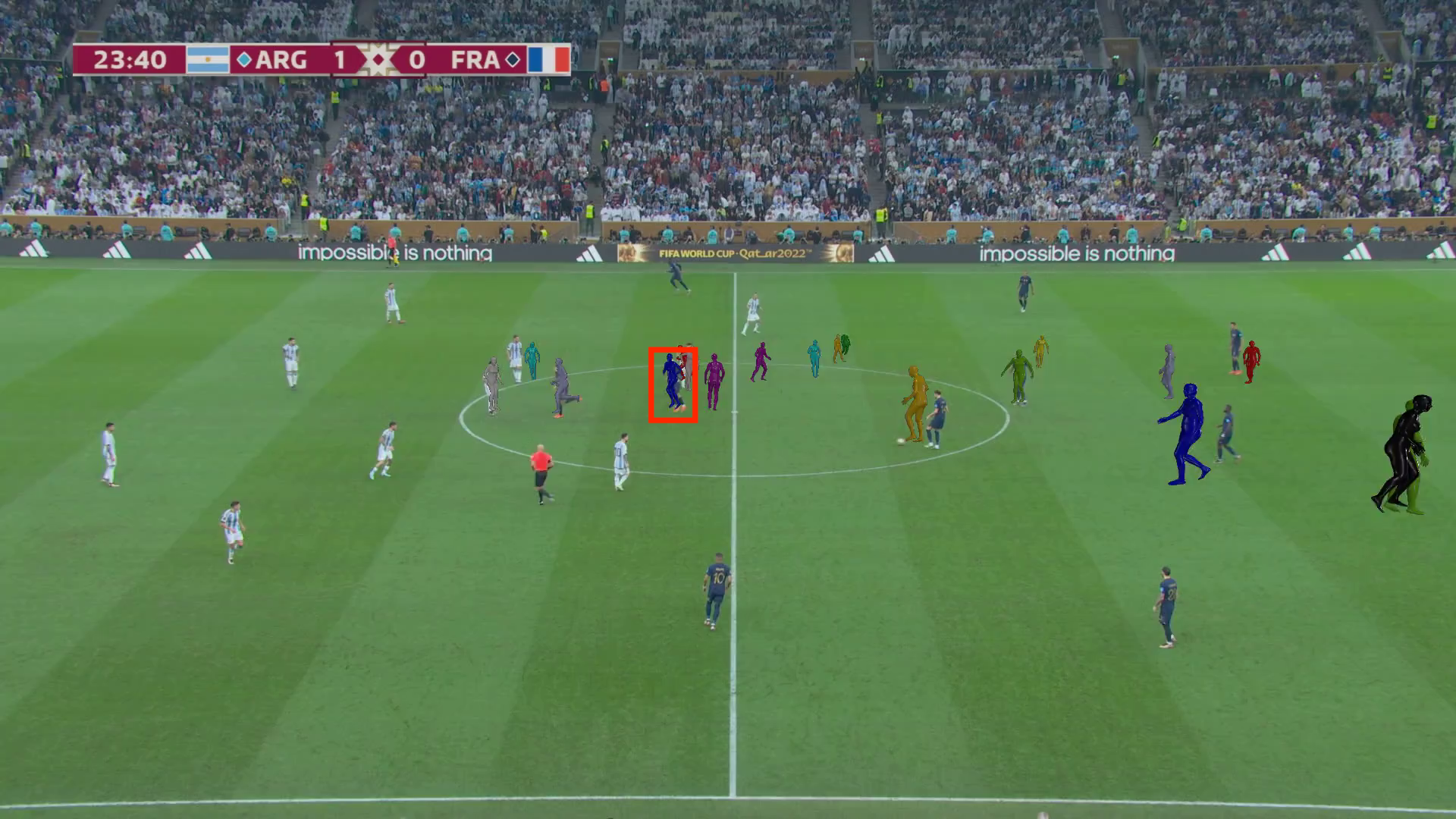}
    \caption{GLAMR}
\end{subfigure}
\caption{Visualization of the GLAMR baseline: We present the results from both Hybrik (with our detections) and GLAMR (using Hybrik as initialization). Despite the good initialization, GLAMR struggles to place SMPL meshes in the correct locations}
\label{fig:supp-glamr}
\end{figure*}

\paragraph{Discussion} In \figref{fig:supp-glamr} we present the results of both HybrIK \cite{Li2021hybrik} and GLAMR. GLAMR utilizes HybrIK to initialize the SMPL estimation. With the provided detections, HybrIK generates accurate SMPL initializations for all subjects in the frame. However, despite the initialization provided by HybrIK, GLAMR struggles to predict plausible trajectories.

Unlike most other SLAM-based methods, GLAMR relies on its learning-based Global Trajectory Predictor to estimate the subjects' trajectories and infer the camera's extrinsic parameters based on these estimated trajectories. However, when the principal axes of the cameras are not parallel to the floor, as in our example, it has difficulty estimating the correct extrinsic parameters, leading to a tendency to place the players on a tilted plane. 

Additionally, we observed that in the implementation of GLAMR, it does not utilize the trajectories of multiple players to improve the estimation of extrinsic parameters (and consequently, the plane). Instead, it solely relies on the trajectory of the player with id=0. Therefore, while GLAMR is able to locate this player (the one annotated with the red rectangle in \figref{fig:supp-glamr}), it fails to accurately place other players, especially those that are far from the reference player. 

\subsection{SLAHMR Baseline}
\paragraph{Implementation Details} For fair comparison with GLAMR, we supply the same detection results used in the GLAMR to SLAHMR \cite{ye2023decoupling}. Additionally, we made a few changes to the official SLAHMR Implementation:
\begin{enumerate}
\item We notice that SLAHMR tends to overlook a few subjects when the number of subjects is relatively large. We made the following changes to the official implementation to address this:
\begin{enumerate}
    \item We increased the constant MAX\_NUM\_TRACKS in the preprocessing code of SLAHMR from 12 to 30. This change allows SLAHMR to keep track of all subjects.
    \item For 4DHuman, we lowered the confidence threshold to 0.5. These changes were made to ensure that all potential players are correctly recognized.
\end{enumerate}
\item We observed that during the motion chunk stage, the optimization failed to converge due to an incorrect floor estimation. Therefore, we specified in the configuration to use a shared floor for all players. In this way, the model will try to align all players to the same floor and yield slightly improved results. Additionally, we enabled the "est\_floor" parameter in the configuration file, allowing the model to estimate the floor normal rather than assuming it is parallel to the xy-plane. We found that this approach improves the performance, particularly when the camera is slightly tilted, as in our case.
\end{enumerate}
Following the original paper, we first run DROID-SLAM\cite{teed2021droid} over the entire video, partition the video into chunks of 100 frames each and optimize each chunk separately. This is because the motion prior model of SLAHMR, HuMoR\cite{rempe2021humor}, is trained on short motion clips and it is recommended by the official HuMoR repository that it should be applied to short clips of 2-3 seconds.

\begin{figure*}[htbp]
\centering
\begin{subfigure}[b]{0.45\linewidth}
    \includegraphics[width=\linewidth]{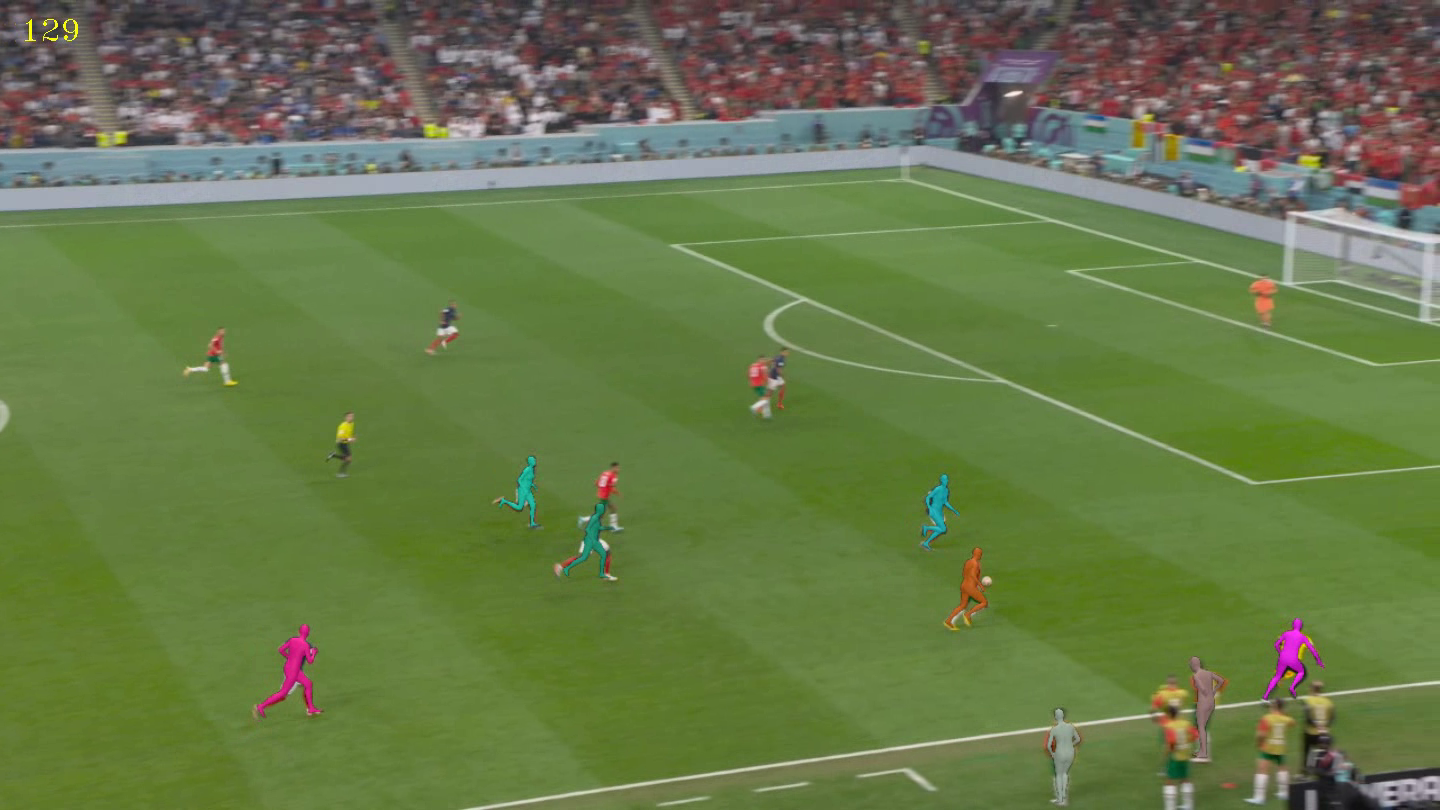}
    \caption{4DHuman (without modification)}
\end{subfigure}
\begin{subfigure}[b]{0.45\linewidth}
    \includegraphics[width=\linewidth]{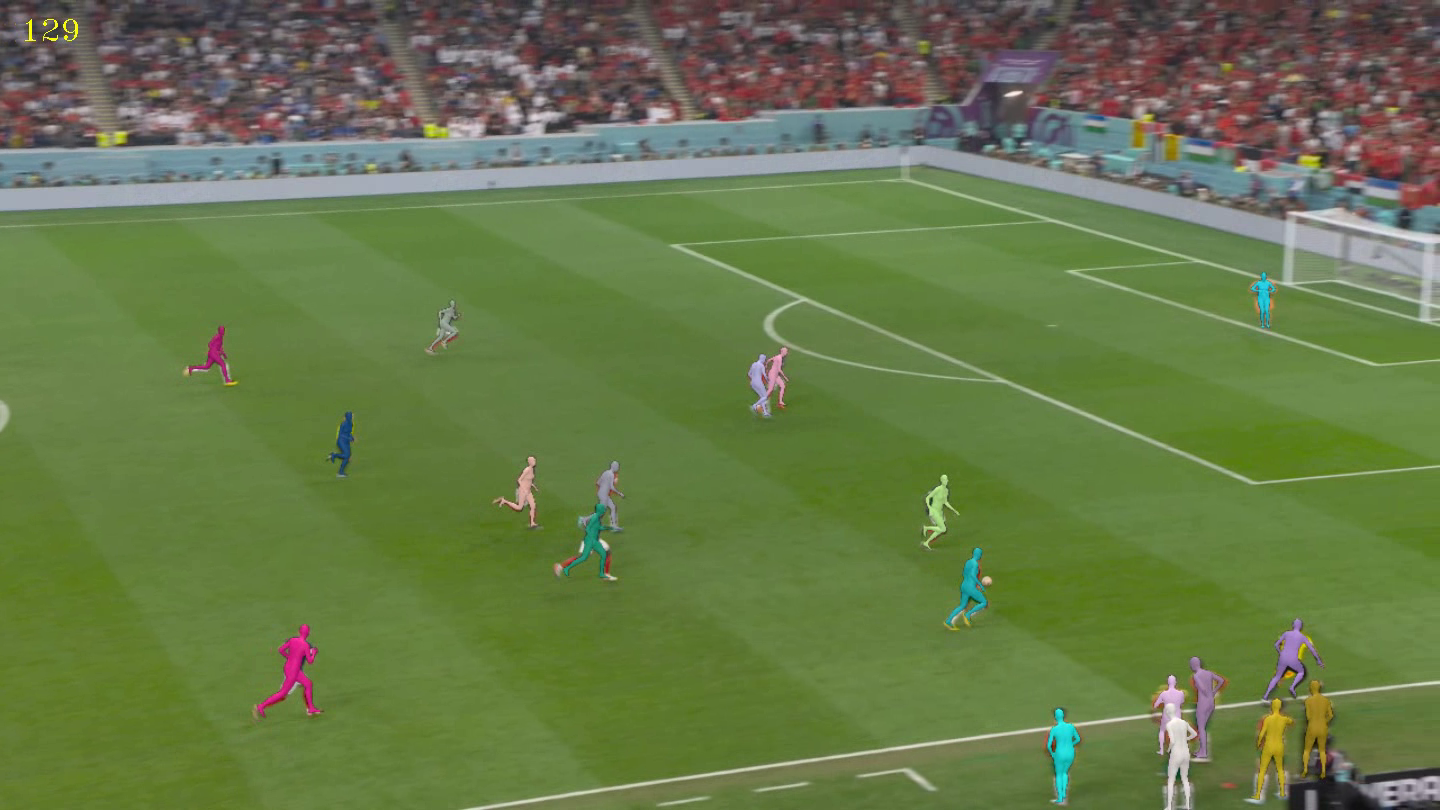}
    \caption{4DHuman (with modification)}
\end{subfigure}

\begin{subfigure}[b]{0.45\linewidth}
    \includegraphics[width=\linewidth]{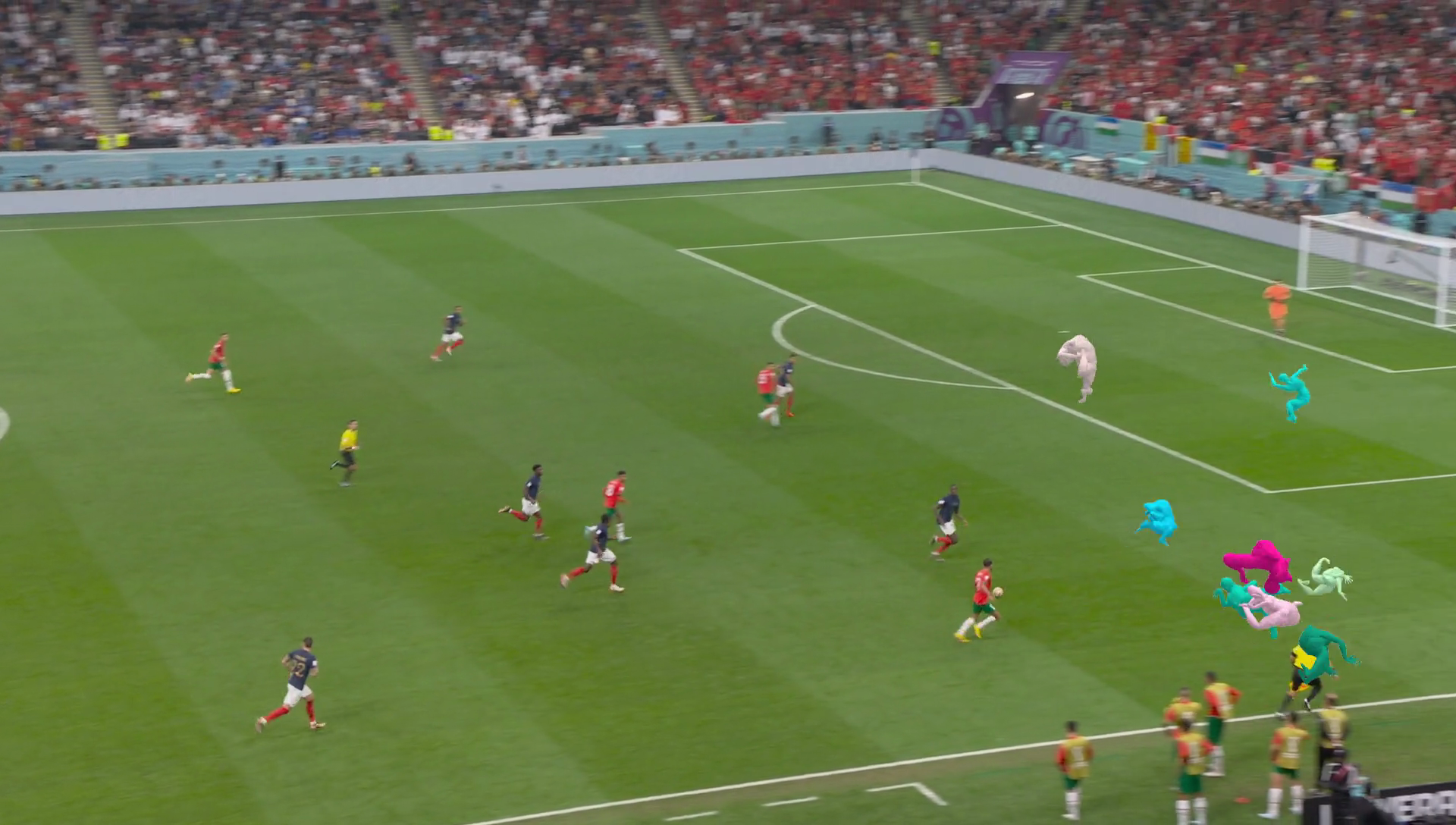}
    \caption{SLAHMR (without modification)}
\end{subfigure}
\begin{subfigure}[b]{0.45\linewidth}
    \includegraphics[width=\linewidth]{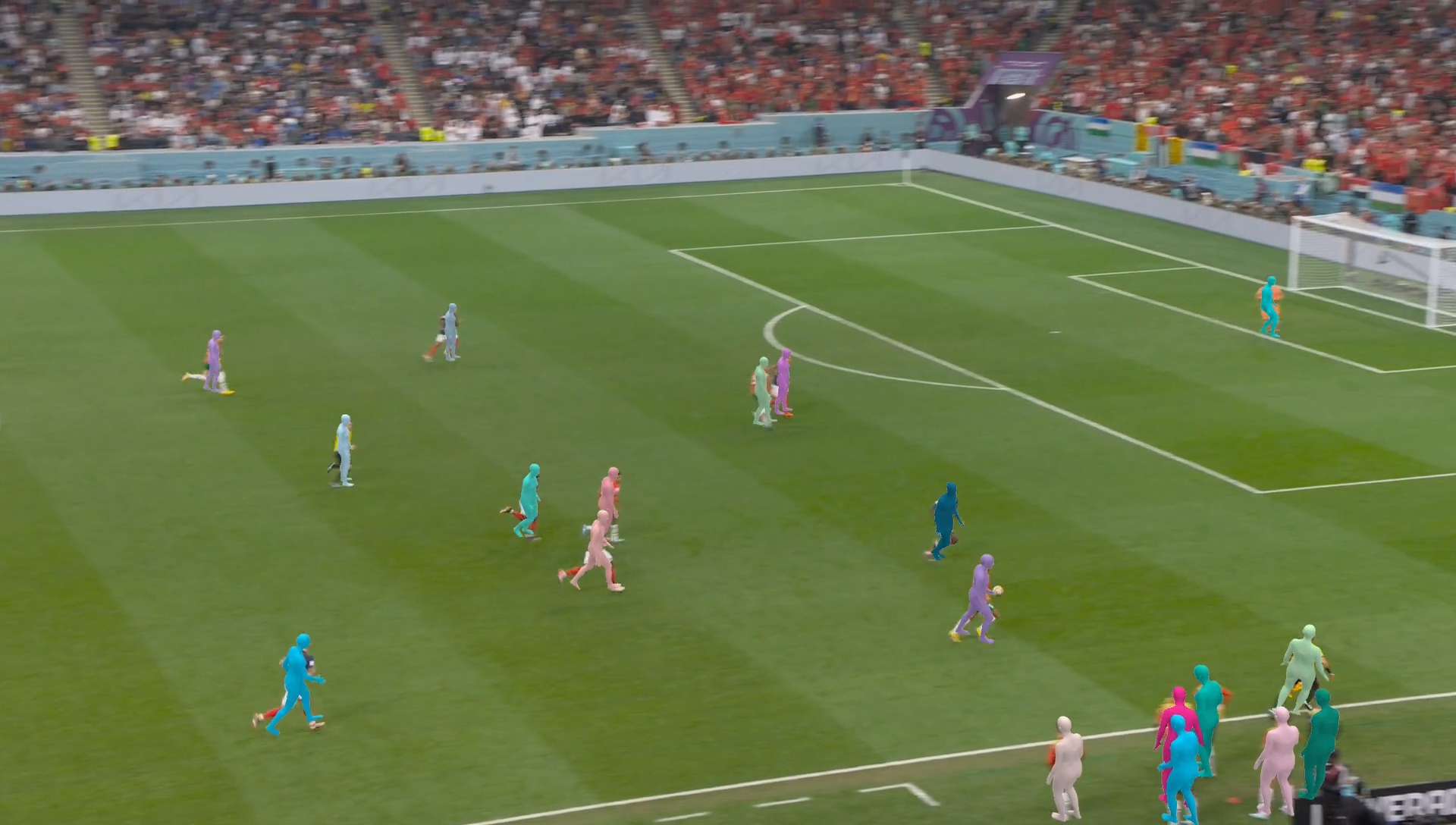}
    \caption{SLAHMR (with modification)}
\end{subfigure}

\begin{subfigure}[b]{0.45\linewidth}
    \includegraphics[width=\linewidth]{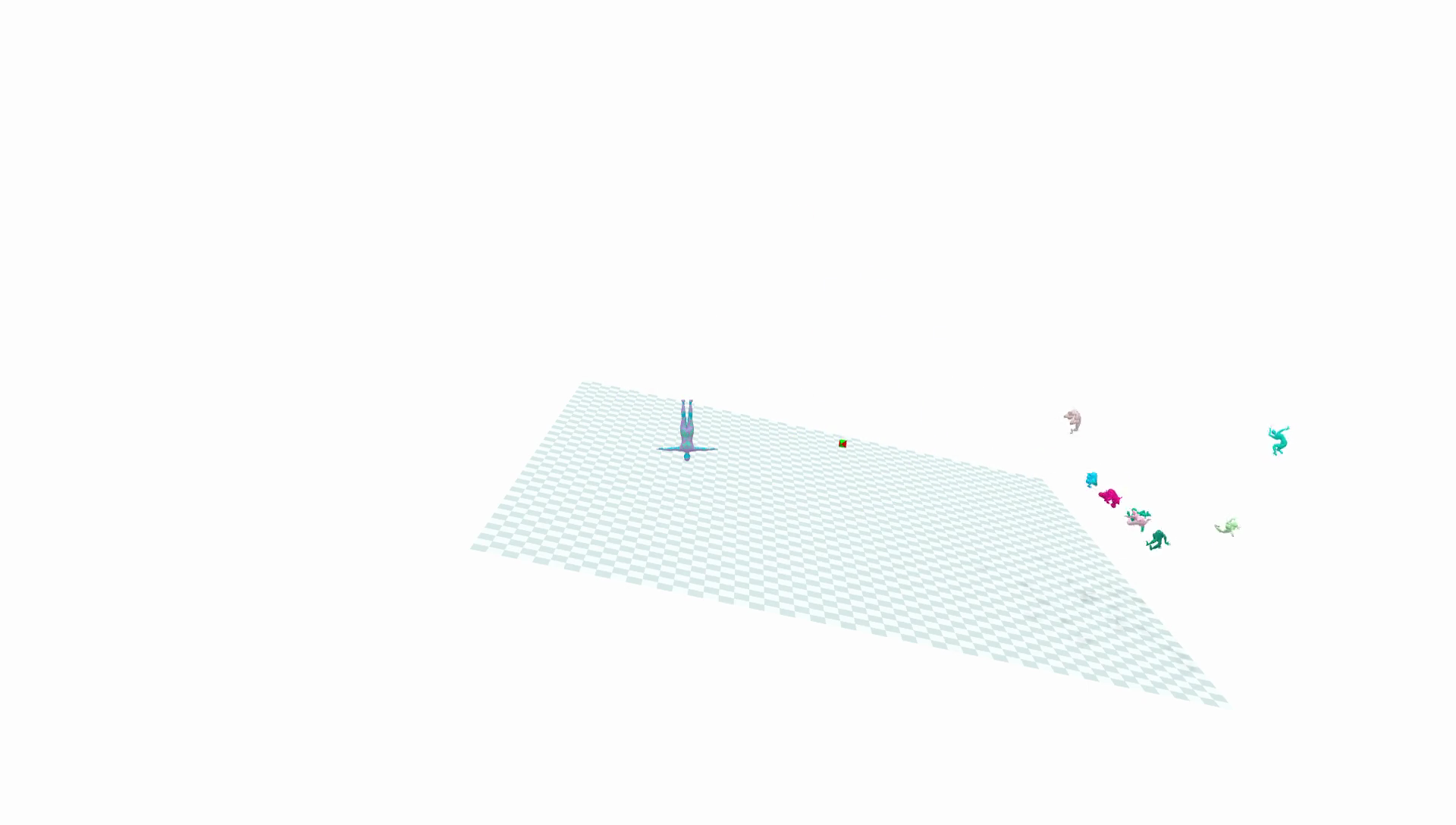}
    \caption{SLAHMR (without modification) in World Coordinate Frame}
\end{subfigure}
\begin{subfigure}[b]{0.45\linewidth}
    \includegraphics[width=\linewidth]{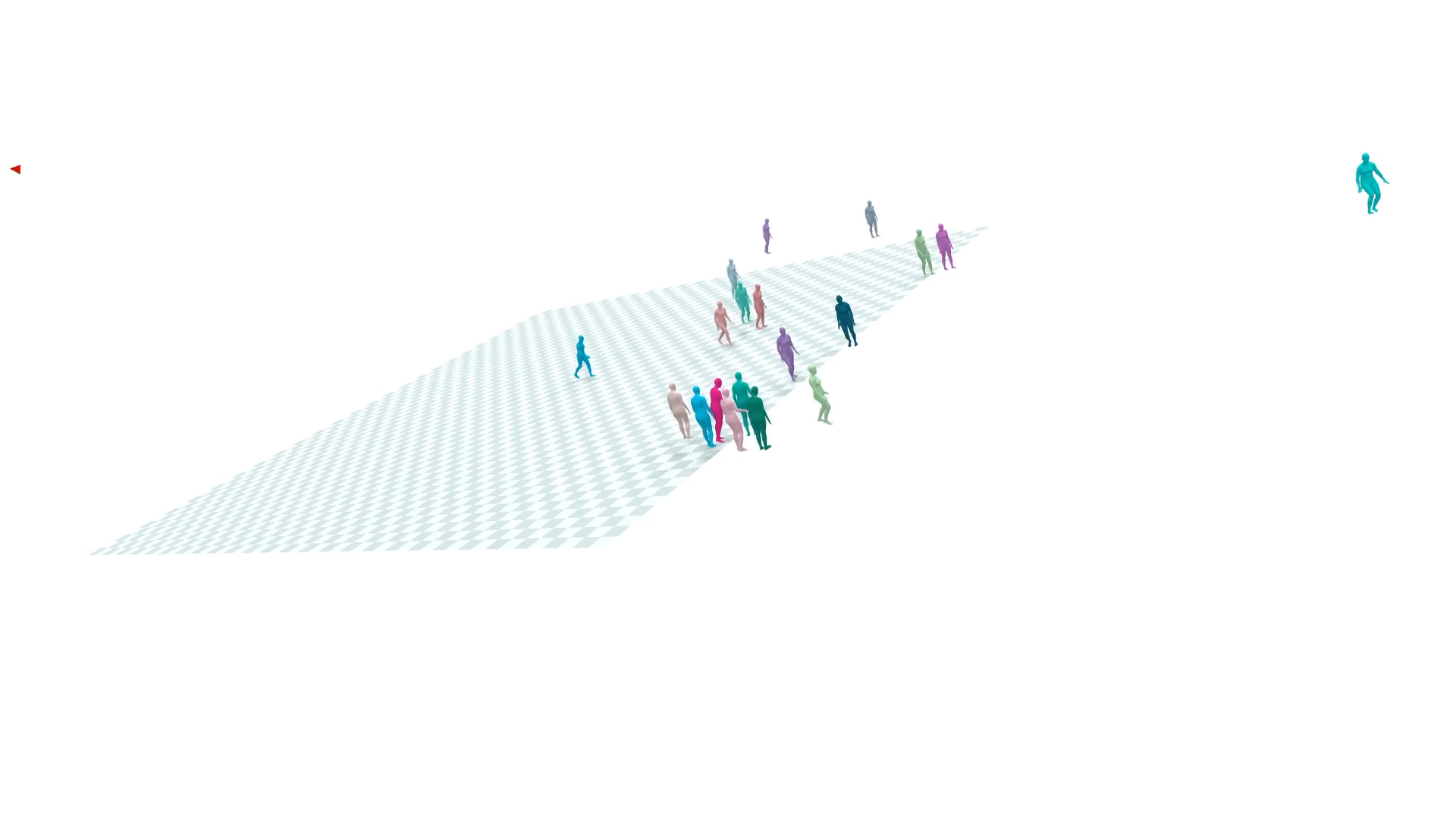}
    \caption{SLAHMR (with modification) in World Coordinate Frame}
\end{subfigure}
\caption{Visualization of SLAHMR and 4DHuman: the left and right columns show the results before and after the modification. While SLAHMR appears to produce seemingly reasonable results (as shown in subfigure f), it does not have the correct scale due to an incorrect focal length. Specifically, the distance between the players should be much larger, as the stadium is approximately 70 meters wide.}
\label{fig:supp-slahmr}
\end{figure*}

\subsubsection{Discussion}
In \figref{fig:supp-slahmr}, we ablate the impacts of our modifications on SLAHMR. With our modification, SLAHMR is able to generate relatively accurate and feasible trajectories in our data. However, while able to produce a reasonable trajectory for individual subjects, SLAHMR struggles to identify the correct relative positioning between different subjects (see  subfigure \figref{fig:supp-slahmr} f).

\begin{figure}
    \centering
    \includegraphics[width=0.8\linewidth]{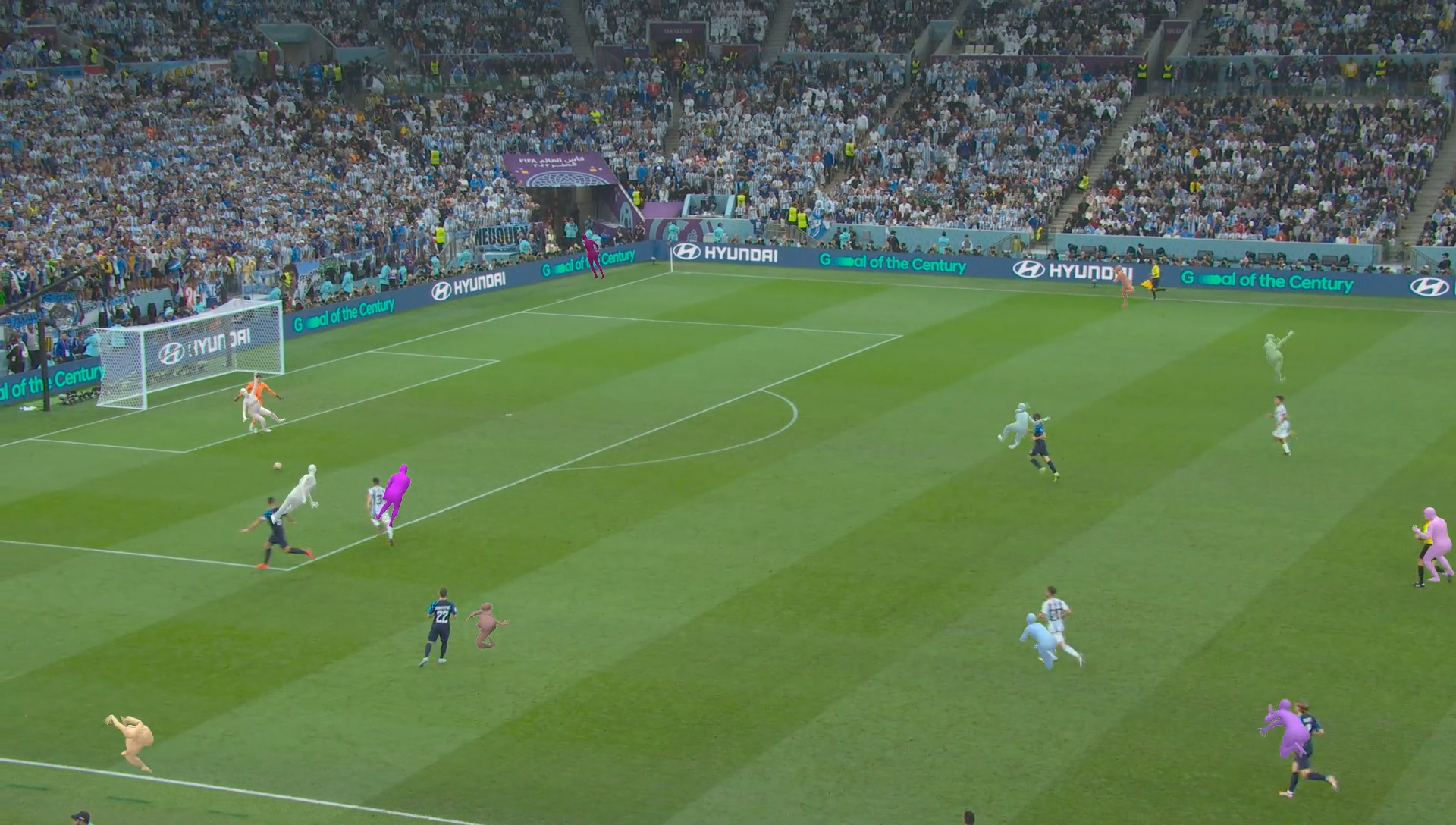}
    \caption{Visualization of typical failure cases of SLAHMR: When the distances between the players are relatively large, SLAHMR struggles to locate the floor, resulting in complete divergence.}
    \label{fig:supp-slahmr-failure-cases}
\end{figure}

While the modification improves the performance of SLAHMR on some sequences, we note that the motion chunk stage remains very fragile and could easily diverge, especially during fast camera movements, which are quite common in broadcasting scenarios. The core issue lies in SLAHMR's need to estimate the floor before introducing the motion prior model. However, the only loss that aligns the players to the floor comes from the motion prior model. This creates a chicken-egg situation: if the players are already roughly on the same plane without the motion prior model, SLAHMR can converge to reasonable results. However, if this is not the case, the motion prior model will not provide any meaningful gradient, leading to complete divergence, as is shown in \figref{fig:supp-slahmr-failure-cases}. Specifically, we observed that for some sequences, SLAHMR fails to converge on as many as half of the chunks, even with ground-truth camera parameters. This can be confirmed from the higher PA-MPJPE loss compared to 4DHuman, as shown in Table 3 of the main paper (which is related to the divergence).

Similar to GLAMR, our evaluation of SLAHMR on \datasetname reveals several limitations: 
\begin{inparaenum}[1)]
\item SLAHMR has a tendency to generate overly smooth motions which poses challenges in capturing fast-paced movements.
\item the motion chunk stage can be quite unstable with large focal length and when players are not standing close to each other,
\item We noticed that although the camera trajectory remains smooth across the boundary of each chunk, there is a visible gap in the predicted SMPL meshes between the chunks. This issue could potentially be mitigated if SLAHMR divides the sequence into overlapping clips, thereby enforcing smoothness regularization across the chunks, similar to the approach employed in PACE \cite{kocabas2024pace}.
\item  the optimization is time-consuming: 40 minutes per 100 frames with 4 subjects as reported in the original paper (which aligns with our observations).
\end{inparaenum}

\subsection{Evaluation}
To align the predicted SMPL poses with the ground-truth, we employed a greedy matching algorithm based on Intersection-over-Union (IoU), comparing 2D bounding boxes of the ground truth with 2D predictions. We found sometimes baselines split trajectories in case of re-entries or lose track, so we merge tracks corresponding to the same ground truth subject during post-processing. For evaluation, we only consider the subjects and frames when they are both available in the prediction and the ground-truth, and the MPJPE is calculated with selected SMPL keypoints (including the nose, neck, shoulders, wrists, elbows, hips, knees, and ankles) which are generally more reliable.

\section{Additional Statistics on \datasetname}
\label{sec:supp_statistics}
\begin{figure}
    \centering
    \includegraphics[width=0.75\linewidth]{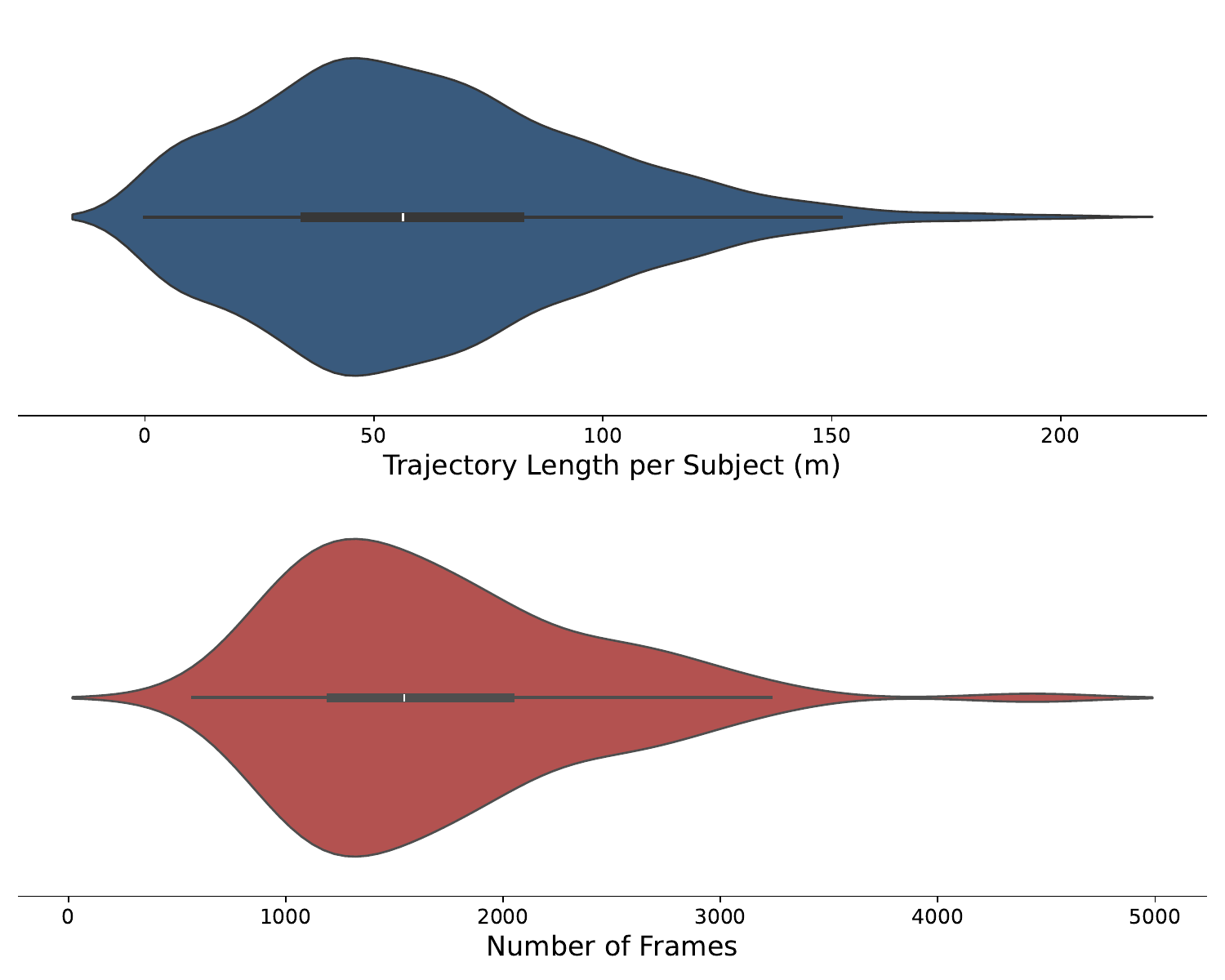}
    \caption{Distribution of player trajectory lengths (top) and clip lengths (bottom) in \datasetname. The availability of long trajectories up to 200 m sets \datasetname apart from existing datasets.}
    \label{fig:supp-stat}
\end{figure}

\begin{figure}
    \centering
    \includegraphics[width=0.45\linewidth]{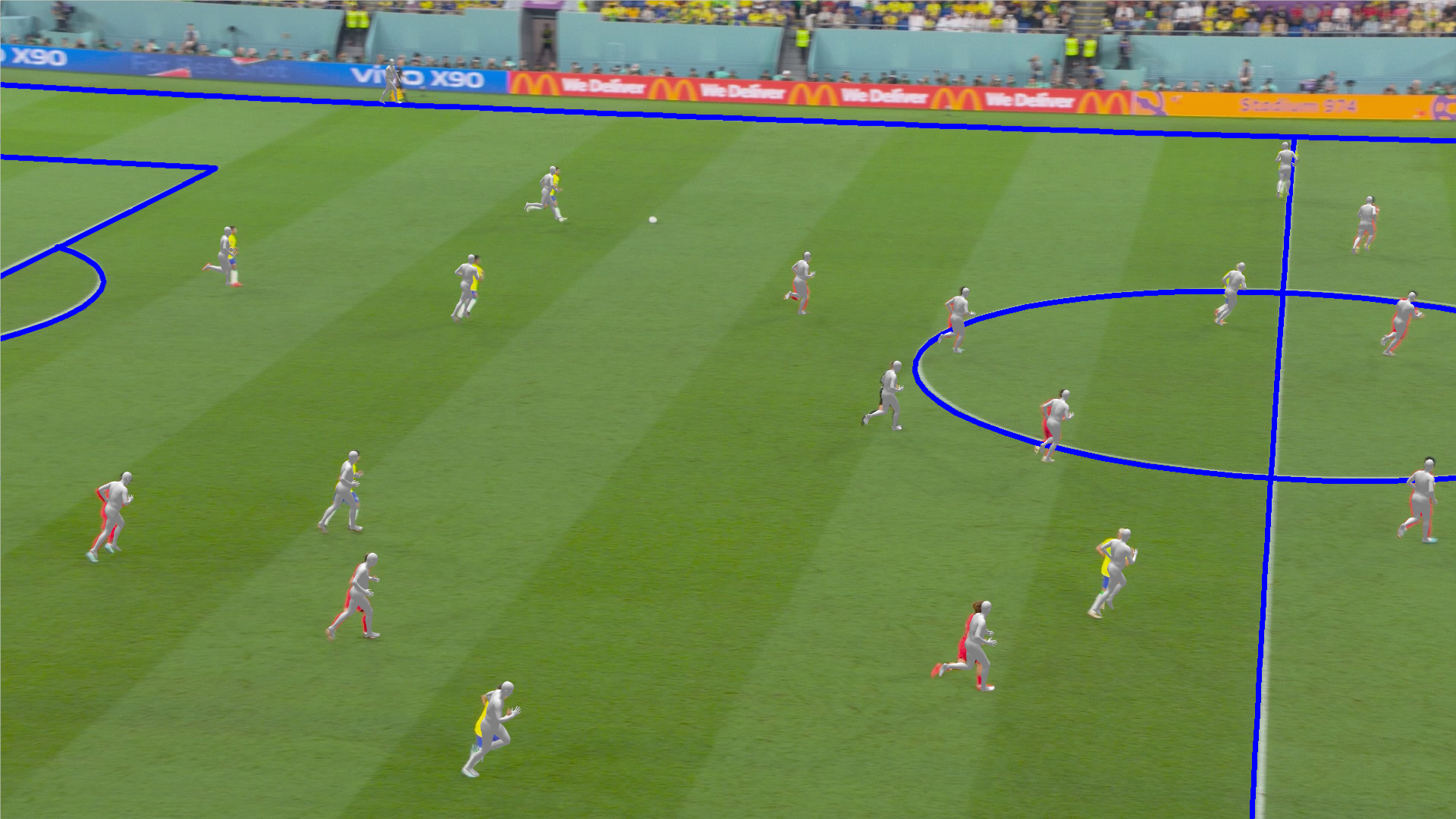}
    \includegraphics[width=0.45\linewidth]{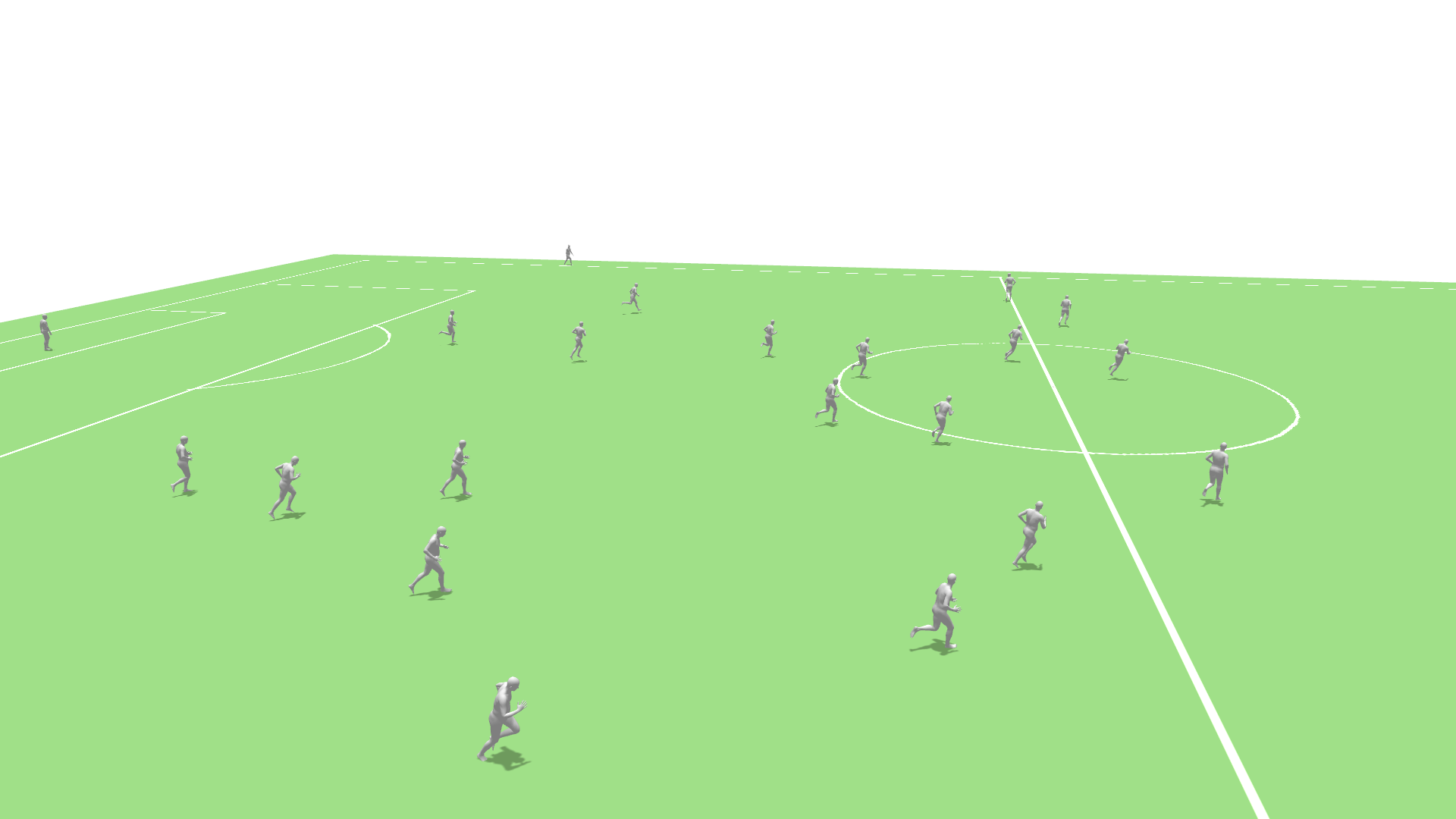}
    \includegraphics[width=0.45\linewidth]{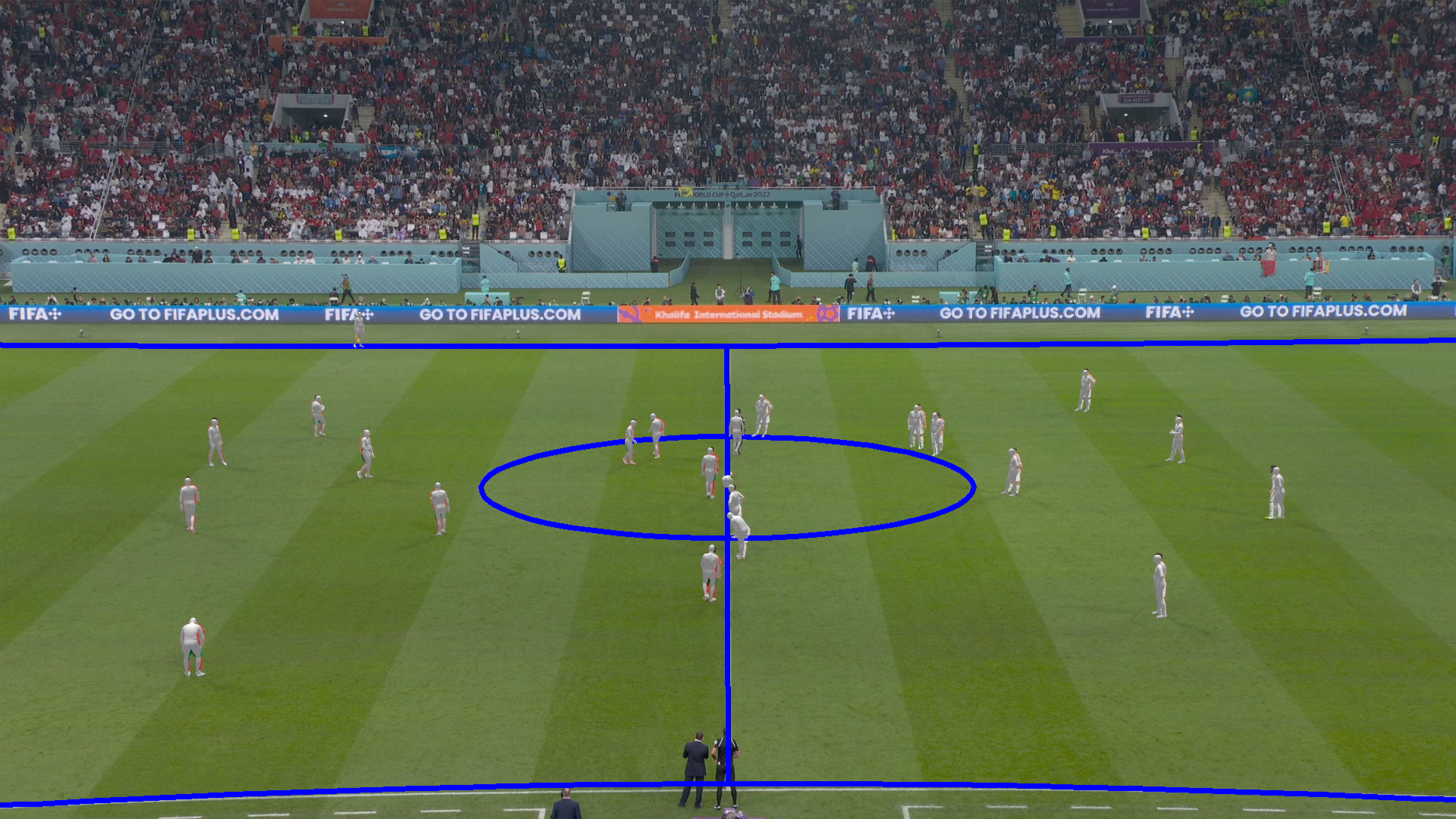}
    \includegraphics[width=0.45\linewidth]{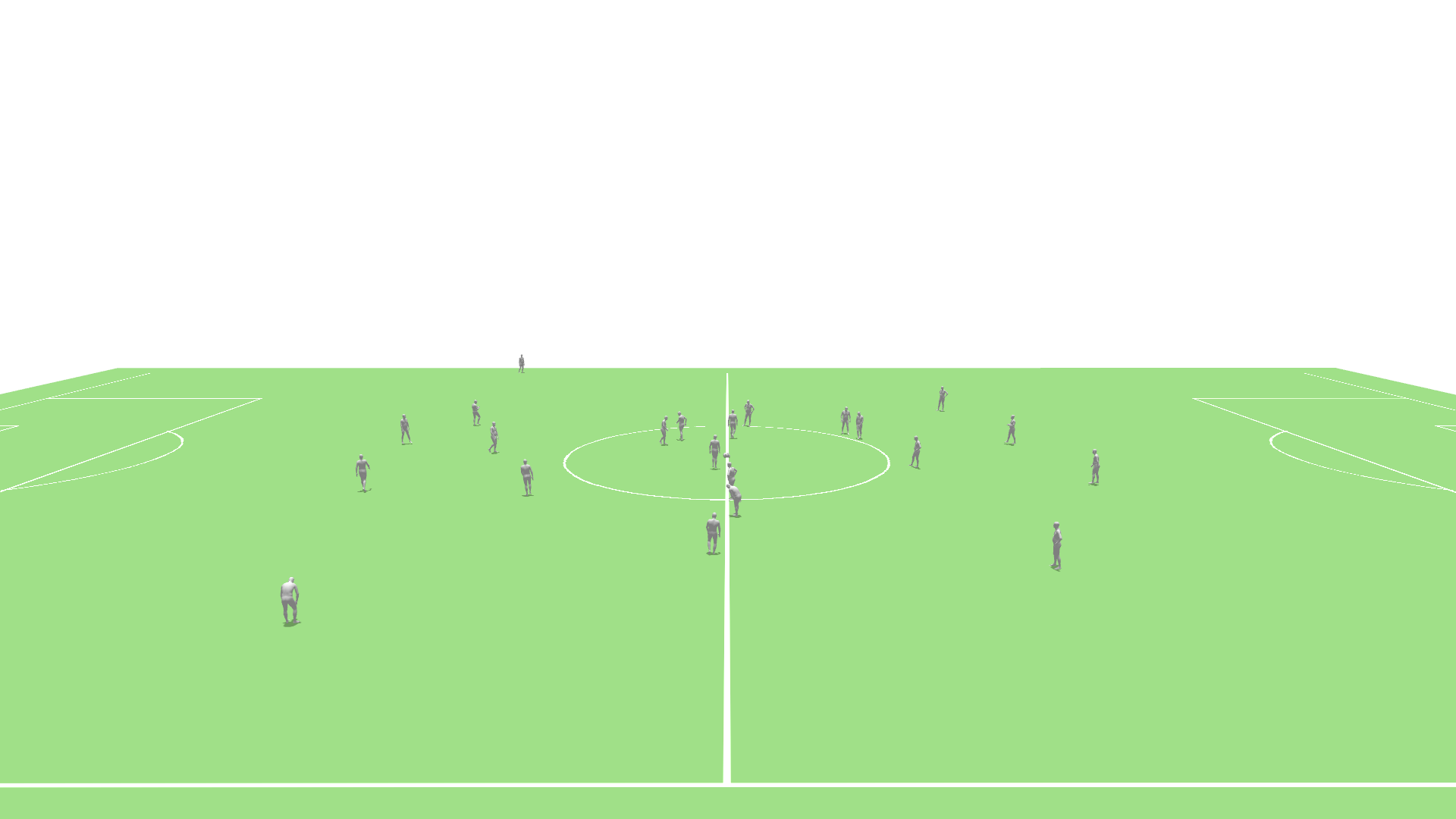}
    \includegraphics[width=0.45\linewidth]{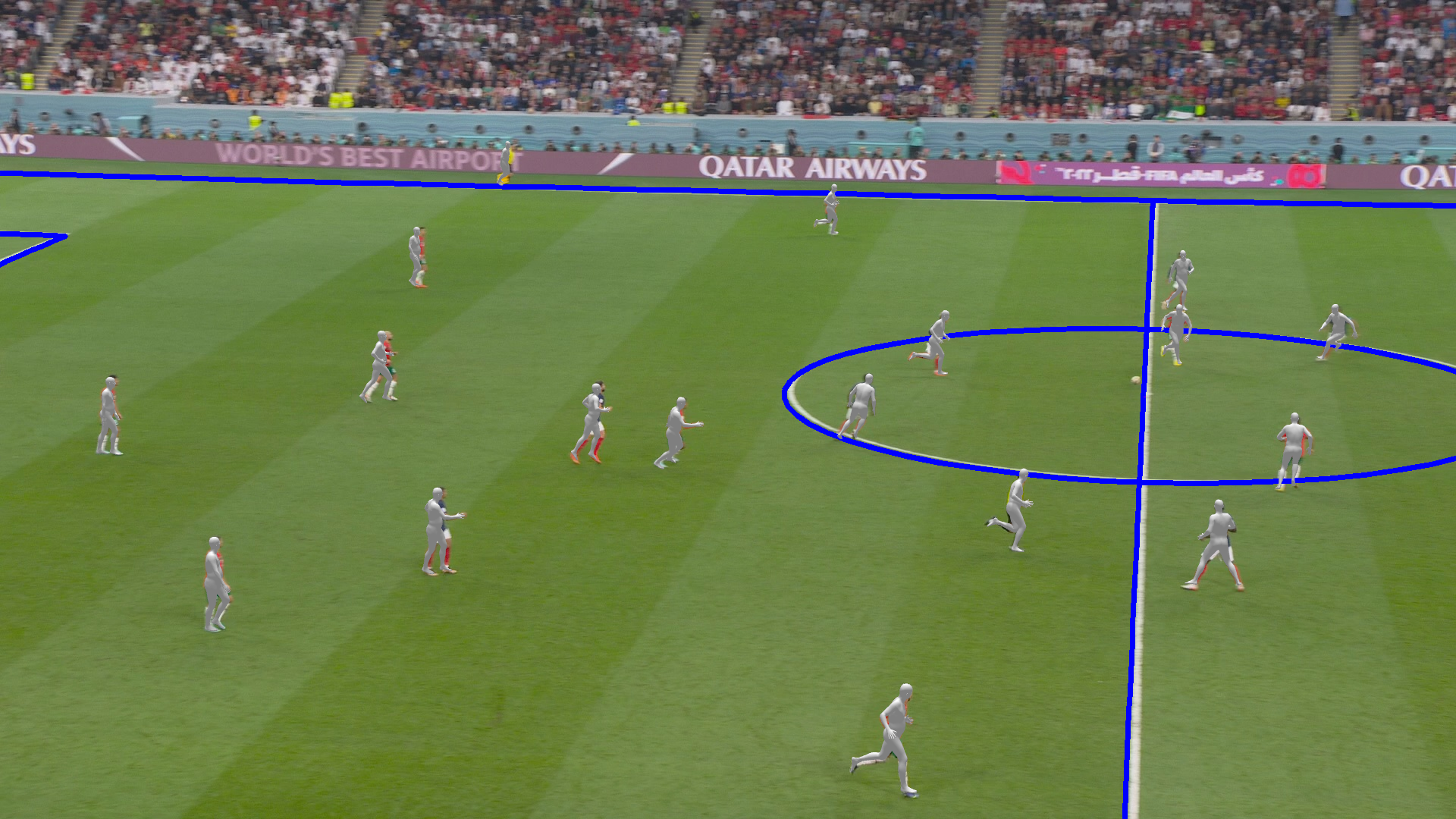}
    \includegraphics[width=0.45\linewidth]{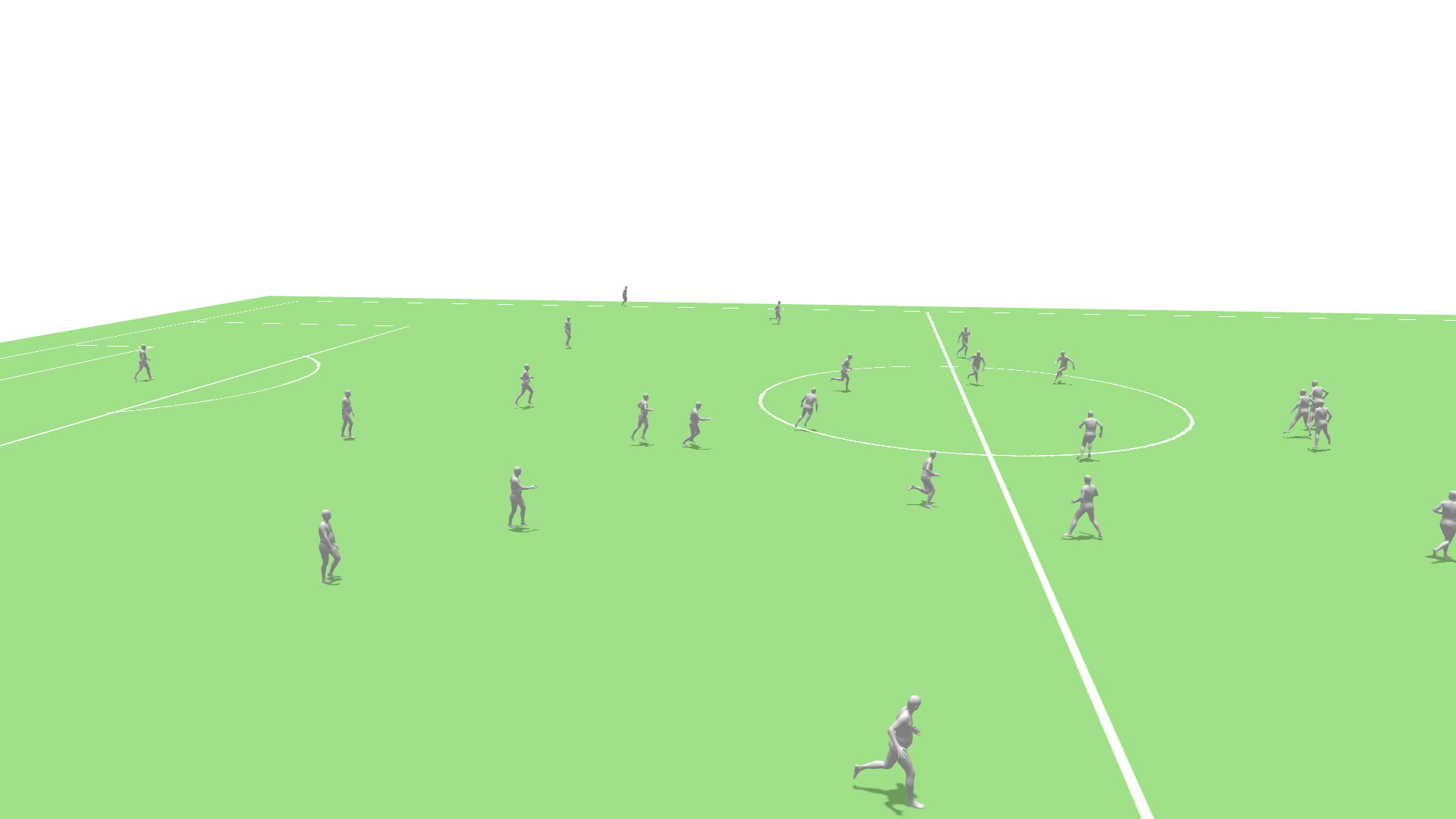}
    \includegraphics[width=0.45\linewidth]{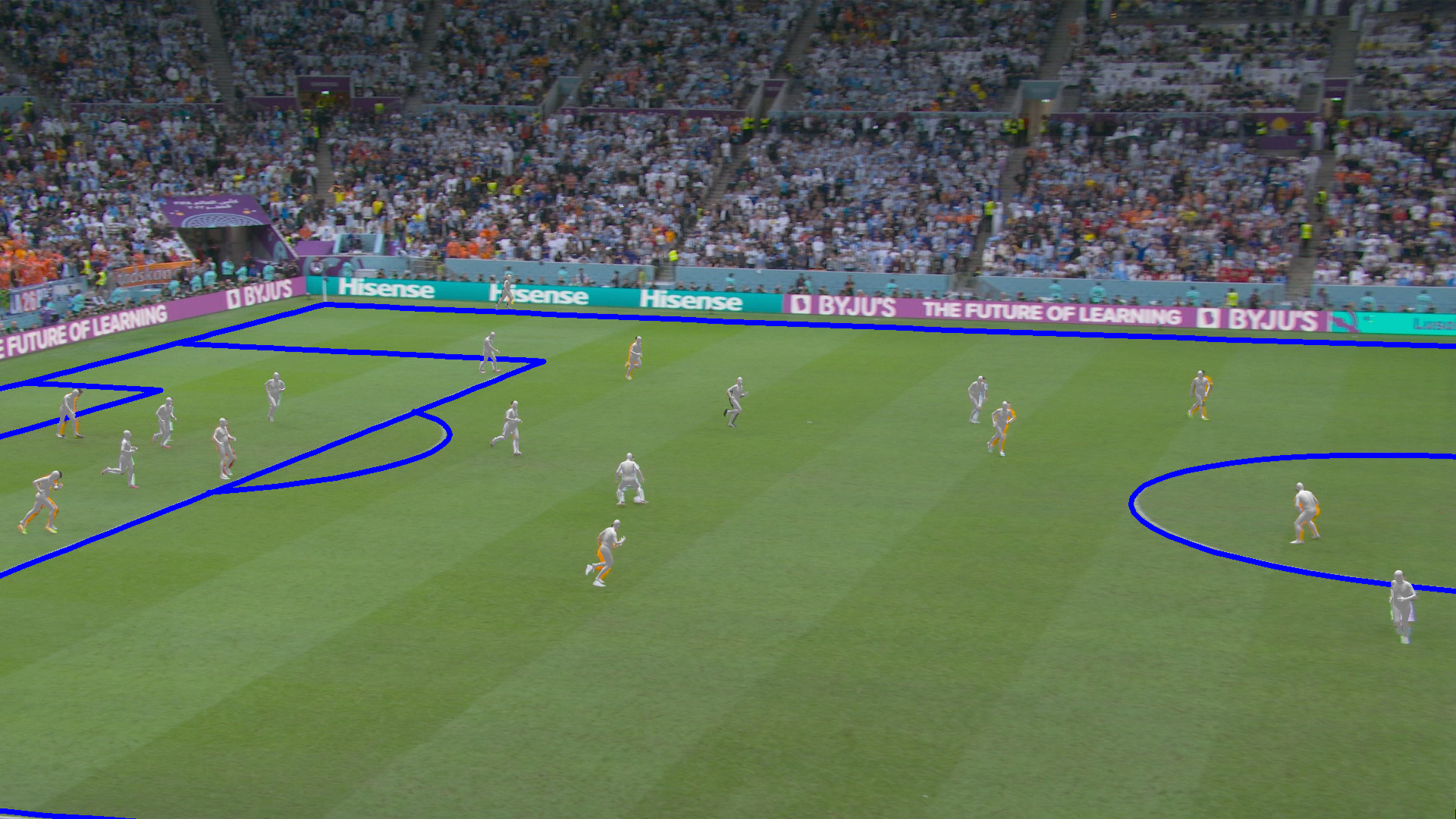}
    \includegraphics[width=0.45\linewidth]{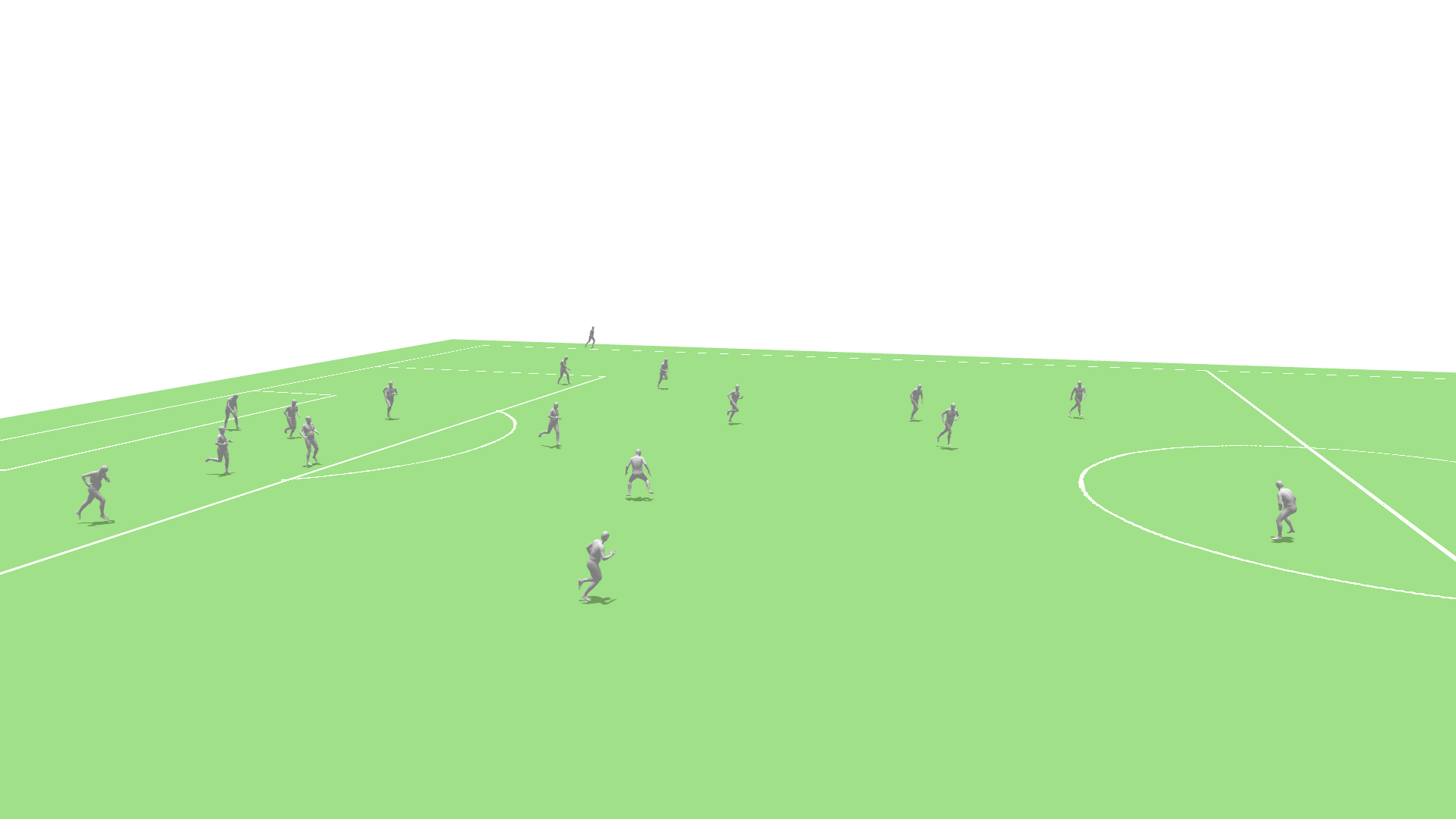}
    \caption{Additional sample images: We include more sample images from our dataset. These images demonstrate that our dataset can provide accurate SMPL meshes and camera parameters, even when the camera zooms in, and fewer corresponding points of field markings are available. }
    \label{fig:supp-more-samples}
\end{figure}

We plot the distribution of sequence lengths and per-player trajectories appearing in \datasetname in \figref{fig:supp-stat}. For additional sample images, please refer to \figref{fig:supp-more-samples}.

\end{document}